%% file: main.tex
\documentclass[journal]{IEEEtran}
\usepackage{graphicx}
\usepackage{graphics} % for resizebox

\usepackage{amssymb}
\usepackage{booktabs}

\usepackage{times}
\usepackage{epsfig}

\usepackage{blindtext}
\usepackage{caption}
\usepackage{amsfonts}
\usepackage[inline]{enumitem}
\usepackage{algorithmic}
\usepackage{textcomp}

\usepackage{multirow}
\usepackage{float}
\usepackage{adjustbox}
\usepackage{etoolbox}
\usepackage{dsfont}

\usepackage{lipsum}
\usepackage{stfloats}
\usepackage{multicol}
\usepackage{amsmath,bm}

\usepackage{array}

\usepackage{tabulary}
\usepackage{xcolor}

\usepackage{paralist}

\usepackage[font=footnotesize]{subfig}
\usepackage{ragged2e}

\usepackage[breaklinks,colorlinks]{hyperref}
\usepackage{bbding}
\definecolor{blue}{rgb}{0,0.5,1}
\hypersetup{colorlinks=true,linkcolor=red,citecolor=blue,urlcolor=magenta}
\usepackage{makecell}
\usepackage{stfloats}

\newcommand{\fn}[1]{\footnotesize{#1}}
\newcommand{\green}[1]{\textcolor[RGB]{96,177,87}{#1}}
\newcommand{\gbf}[1]{\green{\bf{\fn{(#1)}}}}

\definecolor{ZJM}{HTML}{9932CC}
\definecolor{LHY}{HTML}{CD5B45}

\newcommand{\rev}[1]{\textcolor{black}{#1}}

\usepackage{icomma}

\ifCLASSINFOpdf

\else

\fi

\begin{document}

\title{TransKD: Transformer Knowledge Distillation for Efficient Semantic Segmentation}

\author{Ruiping Liu, Kailun Yang\IEEEauthorrefmark{1}, Alina Roitberg, Jiaming Zhang, Kunyu Peng, Huayao Liu,\\Yaonan Wang, and Rainer Stiefelhagen
\IEEEcompsocitemizethanks{
\IEEEcompsocthanksitem This work was supported in part by funding from the pilot program Core-Informatics of the Helmholtz Association (HGF), in part by the National Natural Science Foundation of China (No. 62473139), in part by Hangzhou SurImage Technology Company Ltd, in part by the Ministry of Science, Research and the Arts of Baden-Württemberg (MWK) through the Cooperative Graduate School Accessibility through AI-based Assistive Technology (KATE) under Grant BW6-03, and in part by the Helmholtz Association Initiative and Networking Fund on the HAICORE@KIT and HOREKA@KIT partition.
\IEEEcompsocthanksitem R. Liu, J. Zhang, K. Peng, and R. Stiefelhagen are with the Institute for Anthropomatics and Robotics, Karlsruhe Institute of Technology, 76131 Karlsruhe, Germany.
\IEEEcompsocthanksitem K. Yang and Y. Wang are with the School of Robotics and the National Engineering Laboratory of Robot Visual Perception and Control Technology, Hunan University, Changsha 410082, China.
\IEEEcompsocthanksitem A. Roitberg is with the Institute for Artificial Intelligence, the University of Stuttgart, 70569 Stuttgart, Germany.
\IEEEcompsocthanksitem J. Zhang is also with the Institute for Visual Computing, ETH Zurich, 8092 Zurich, Switzerland.
\IEEEcompsocthanksitem H. Liu is with NIO, Shanghai 201804, China.
\IEEEcompsocthanksitem \IEEEauthorrefmark{1}Corresponding author (E-Mail: kailun.yang@hnu.edu.cn.)
}
}

% The paper headers
\markboth{IEEE Transactions on Intelligent Transportation Systems, September~2024}%
{Liu \MakeLowercase{\textit{et al.}}: TransKD}

% make the title area
\maketitle

\begin{abstract}
\input{Tex_content/abstract}
\end{abstract}

% Note that keywords are not normally used for peerreview papers.

\IEEEpeerreviewmaketitle

\section{Introduction}\label{sec:introduction}

\input{Tex_content/introduction}

\section{Related Work}
\input{Tex_content/related_work}

\section{Proposed Framework: TransKD}

\subsection{Overview}

At the heart of this work is the transformer-to-transformer knowledge transfer for efficient semantic segmentation.
To strike a better balance between accuracy, computational costs, and the required amount of pre-training, we for the first time look at semantic segmentation through the lens of \textit{multi-source knowledge distillation} within visual transformers. 
While transformer-to-transformer knowledge distillation is explored in natural language processing~\cite{jiao2019tinybert,dong2021efficientbert,wang2020minilm}, 
it has been rather overlooked in semantic segmentation and vision transformers in general, where most of the models still center around CNN-based knowledge distillation.
Adopting such methods to suit visual transformers is not trivial due to diverging architecture-specific building blocks arising, \textit{e.g.}, from the patchifying process.

Existing semantic segmentation transformers generally follow either an isotropic structure~\cite{zheng2021setr, zhou2022fan,chen2022vision_transformer_adapter} or a four-stage hierarchical structure~\cite{wang2022pvt_v2,xie2021segformer,yang2022lite_vision_transformer, wang2021pvt, lee2021mpvit}. 
Since the latter one is better at modeling multi-scale long-range information~\cite{liu2021swin, wang2021pvt}, it proves more versatile in dense prediction tasks such as pixel-wise semantic segmentation.
Thereby, in this work, knowledge distillation is carried out on four-stage transformers, as illustrated in Fig.~\ref{fig:KD_structure}.

The first step of the general visual transformer pipeline is to ``patchify'' the input $\mathbf{I}$ (an image or a feature map):
\begin{equation}\label{eq:patch_embd}
    \mathbf{E} = PatchEmbed\left(\mathbf{I}\right).
\end{equation}
This ``patchification'' is the key ingredient of vision transformers as it allows us to cast vision-based tasks as sequence-to-sequence problems for which this type of models was initially built~\cite{vaswani2017attention}.
The patch embedding module splits the input images ohes. \rev{Here, $H$, $W$, and $C$ represent the height, the width, and the number of channels of the input image or feature map of the embedding module, respectively. The module then} transforms these patches (mostly via a convolution operation) into \rev{a sequence of} representative patch embeddings $\mathbf{E} \in \mathbb{R}^{N\times C^{\prime}}$, where $N$ is the number of patches and $C^{\prime}$ the number of the channels.
As often highlighted in previous research, the resulting patch embeddings convey important information about long-range sequential relationships~\cite{zheng2021setr,dong2022cswin}, location priors~\cite{ zheng2021setr, dosovitskiy2021vit, zhou2022fan}, and local continuity cues~\cite{wang2022pvt_v2, xie2021segformer,lee2021mpvit}. 
\rev{Both global context and local information contained in patch embeddings are relevant to semantic segmentation.}

Next, the patch embeddings are passed to a stack of transformer encoding layers consisting of a multi-head self-attention and a position-wise fully connected feed-forward network~\cite{dosovitskiy2021vit,vaswani2017attention}.
The outputs of these  layers are then reshaped into a feature map $\mathbf{F}$: 
\begin{equation}\label{eq:transformer_encod}
    \mathbf{F} = TransformerEncode\left(\mathbf{E}\right),
\end{equation}
where the feature map $\mathbf{F} \in \mathbb{R}^{{C^{\prime}\times{H^{\prime}\times{W^{\prime}}}}}$ marks the output of the current transformer stage and therefore the input to the next layer.
A positive consequence of self-attention exploring all the patch tokens with no restriction to a certain receptive field is the \textit{global} context also being covered in the resulting feature map.
Distilling knowledge from both, feature maps and patch embeddings, allows us to track complementary information within the transformer, such as spatial and sequential relationships. 

Motivated by this, we propose \emph{Transformer-based Knowledge Distillation (TransKD)} -- a novel transformer-to-transformer knowledge distillation framework with the overall goal of building small yet accurate semantic segmentation transformers less reliant on time-consuming pre-training. 
TransKD centers around two types of knowledge transfer: (1) patch embedding distillation and (2) feature map distillation.
The \textit{patch embedding distillation} is performed at each of the four transformer stages, while the feature map distillation leverages the relation-based knowledge distillation to gather information from both within-stage and cross-stage feature maps. An overview of this core pipeline is depicted in Fig.~\ref{fig:KD_structure}(a).

The distillation losses obtained from each stage are added to the original cross-entropy loss $L_{CE}$ during training.
 The overall KD loss then becomes:
\begin{equation}\label{eq:loss}
L = L_{CE} + {\sum_{m=1}^M}{\alpha_m}L_{embd}^m + {\sum_{m=1}^M}{\beta_m}L_{fm}^m,
\end{equation}
where $L_{embd}^m$ and $L_{fm}^m$ refer to the patch embedding loss and the feature map loss between the $m$-th stage of the teacher and the student. As previously mentioned, we explore the typical four-stage transformer in this work, \textit{i.e.}, $M=4$ in our case. The variables $\alpha_m$ and $\beta_m$ are the $m$-th elements of $\mathbf{\alpha}$ and $\mathbf{\beta}$, which control the weights of the patch embedding loss and the feature map loss at each stage, respectively. The vectors $\mathbf{\alpha}$ and $\mathbf{\beta}$ are set as $\left[0.1, 0.1, 0.5, 1\right]$ and $\left[1, 1, 1, 1\right]$.
In the following method descriptions, matrices with superscripts $in$ and $out$ refer to inputs and outputs of the proposed modules, while those with the superscript $mid$ refer to matrices processed within the modules.

In Sec.~\ref{sec:patch_embedding_distillaton}, we present the design of our patch embedding distillation scheme, and in Sec.~\ref{sec:feature_map_distillaton}, we elaborate our feature map distillation method. Finally, in Sec.~\ref{sec:configuration}, we describe the configurations of different variants of the TransKD framework.

\subsection{Patch Embedding Distillation}
\label{sec:patch_embedding_distillaton}
\rev{As previously discussed, the patch embedding incorporates positional information that enhances the feature maps.} To maximize the potential of transformer-to-transformer knowledge distillation, we consider that transformer-specific patch embeddings should be fully exploited.
\rev{Despite the specialized designs, the cumbersome teacher models consistently possess more transformer blocks for feature extraction and additional channels for knowledge retention. To align dimensions between student and teacher patch embeddings for loss function calculation and to introduce trainable parameters for learning channel-wise knowledge, we propose a fundamental module, \emph{Patch Embedding Alignment (PEA)}, to facilitate the simplest patch embedding distillation.} 
Additionally, we design the \emph{Global-Local Context Mixer (GL-Mixer)} and the \emph{Embedding Assistant (EA)} to optimize patch embedding distillation.

\noindent\textbf{Patch embedding alignment.}
We conduct the Patch Embedding Alignment (PEA) through a linear projection for dimensionality transformation. Subsequently, the \rev{Mean Squared Error (MSE)} loss between the student and teacher patch embeddings is calculated, as depicted in Eq.~(\ref{eq:embd}):
\begin{equation}\label{eq:embd}
L_{embd}^m = MSE({\mathbf{{\bm{E}}}}^S_m{{\mathbf{\bm{W}}}_e},\mathbf{{\bm{E}}}^T_m),
\end{equation}
where $\bm{E}_m^S$ and $\bm{E}_m^T$ are \rev{the sequence} of patch embeddings at the $m$-th student and teacher stages. A learnable matrix \rev{$\bm{W}_e \in {\mathbb{R}}^{C^S_m {\times} C^T_m}$} is applied to counter the dimensionality mismatch and \rev{transfer channel-wise knowledge from teacher patch embeddings to the student. $C^S_m$ and $C^T_m$ denote the number of channels in teacher and student models, respectively, at $m$-th stage.} 
PEA bridges the teacher and student transformers and facilitates multi-stage patch embedding streams, while adding only marginal computational complexity.

\begin{figure}[t!]
    \centering
    \includegraphics[width=0.35\textwidth]{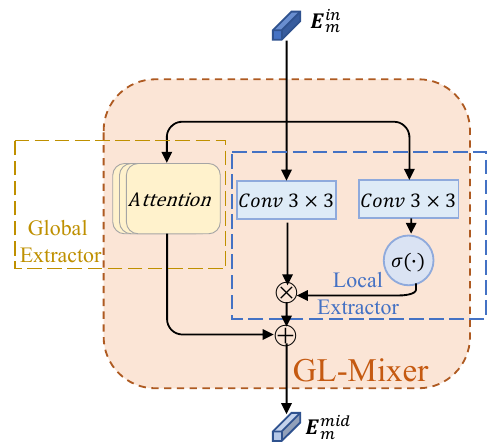}
    \caption{Two-branch architecture of Global-Local Context Mixer (GL-Mixer) in Fig.~\ref{fig:KD_structure}(b). The global context of an embedding is extracted by multi-head attention, whereas the local features are extracted by convolution operations. }
    \label{fig:GL-Mixer}
\end{figure}

\noindent\textbf{Global-local context mixer.}
Patch embeddings deliver heterogeneous patterns~\cite{dosovitskiy2021vit,xie2021segformer}.
Therefore, the global context cues and the fine-grained local features, both essential for semantic segmentation, cannot be fully distilled using linear projection alone.
To our knowledge, we are the first to introduce the two-branch design to the field of knowledge distillation in semantic segmentation. 
The global context is extracted via multi-head attention (left branch in Fig.~\ref{fig:GL-Mixer}), and the local information is extracted via a convolution layer (right branch in Fig.~\ref{fig:GL-Mixer}).
Another convolution operation, followed with sigmoid function $\sigma\left(\cdot\right)$, is used to control the information flow.
Given 
\rev{the sequence of patch embedding at the $m$-th stage} $\mathbf{E}_m^{in}$, the two-branch processing of global and local cues inside the GL-Mixer is formalized in Eq.~(\ref{eq:glmixer}):
\begin{equation}\label{eq:glmixer}
\begin{aligned}
\textit{GL-Mixer}\left(\mathbf{E}_m^{in}\right) &= MultiHeadAttn(\mathbf{E}_m^{in})\\ &+\left(\mathbf{E}_m^{in}*\mathbf{W}+\mathbf{a}\right)\otimes\left(\sigma\left(\mathbf{E}_m^{in}*\mathbf{V}+\mathbf{b}\right)\right),
\end{aligned}
\end{equation}
where $\mathbf{W}$, $\mathbf{V}$, $\mathbf{a}$, and $\mathbf{b}$ refer to learnable parameters for the convolution operations.
As shown in Fig.~\ref{fig:GL-Mixer}, the convolution operations are implemented via $3{\times}3$ convolutions, and the entire GL-Mixer is designed to be lightweight. This way, given a \rev{sequence of} representative patch embedding within a dense prediction transformer, both the semantically rich global context and the detail-rich local features can be harvested for distillation.

\begin{figure}
    \centering
    \includegraphics[scale=0.65]{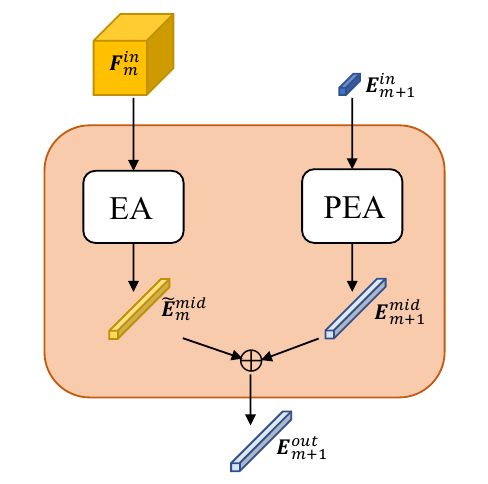}
    \caption{Combination of Embedding Assistant (EA) and Patch Embedding Alignment (PEA) in Fig.~\ref{fig:KD_structure}(c). The knowledge distillation pipeline through the EA modules and the student transformer blocks can be regarded as a medium-sized pseudo assistant model, providing supplementary distillation streams. $\mathbf{F}_m^{in}$ denotes the resulting feature map of the $m$-th stage, which is processed by the student's embedding module to output the patch embedding of the next stage $\mathbf{E}_{m+1}^{in}$.}
    \label{fig:EA_PEA}
\end{figure}

\noindent\textbf{Embedding assistant.}
Knowledge distillation runs into problems when the size gap between student and teacher is significant~\cite{mirzadeh2020teacher_assistant,huang2022knowledge_distillation_stronger_teacher}.
If the teacher becomes very complex, the student eventually lacks the sufficient capacity or mechanics to mimic its behavior despite receiving multi-stage hints.
Mirzadeh~\textit{et al.}~\cite{mirzadeh2020teacher_assistant} introduce a Teacher Assistant (TA) of an intermediate size to bridge the size gap.
TA acts as a mediator between the teacher and the student: it is distilled from the teacher model and then passes the knowledge to the student.
TA follows the same structure as the student and the teacher, except for the numbers of hidden states and transformer layers~\cite{wang2020minilm}.
The number of hidden states is adopted from the student, while the amount of the transformer layers corresponds to the one of the teacher.
\rev{However, implementing Teacher Assistant for knowledge distillation is a two-step process. An additional assistant model TA should be trained, which is larger than the student model, thus introducing extra computation overhead and learning time.}

We draw inspiration from the concept of TA but address the aforementioned drawback into account and propose the \emph{Embedding Assistant (EA)} module which serves as a mediator between the teacher and the student \textit{without} requiring two-stage distillation.
An overview of the proposed EA module is given in Fig.~\ref{fig:KD_structure}(c).
EA is an embedding module whose structure is identical to the one of the student's and the teacher's embedding module and with the number of channels $C$ corresponding to the one of the teacher.
With the same kernel size of the convolution operation to realize patch embedding, the outputs of EA, PEA, and the teacher's embedding module have the same number of patches.
As shown in Fig.~\ref{fig:KD_structure}(c) and Fig.~\ref{fig:EA_PEA}, the results of EA and PEA are fused via element-wise addition stage-by-stage. 

Given that a transformer block of the teacher consists of $L$-transformer layers with an embedding size $d_e$, and a transformer block of the student consists of $M$-transformer layers with an embedding size of $d_e^{\prime}$, the combination of EA modules and the student's transformer blocks can be viewed as a pseudo assistant model with a transformer block of $M$-transformer layers and an embedding size of $d_e$.
Therefore, the medium-sized pseudo assistant model, \textit{i.e.}, the combination of all four stage EAs and student transformer blocks, helps to bridge the student-teacher size gap and provides supplementary information.
Since the embedding method of different models varies greatly, the knowledge distillation framework using the proposed EAs does not achieve full plug-and-play flexibility.
Yet, compared with the teacher assistant~\cite{mirzadeh2020teacher_assistant}, we strengthen the student-teacher distillation without a costly two-step knowledge distillation process.

\subsection{Feature Map Distillation}
\label{sec:feature_map_distillaton}
Many knowledge distillation methods deploy a single-stage feature-based scheme for semantic segmentation~\cite{shu2021channel_distillation,wang2020intra_class_feature_variation_distillation,ji2022structural_statistical_texture_distillation}, \textit{i.e.}, information is exchanged at a single network depth.
Our next consideration is that \textit{cross-stage} relation-based knowledge distillation is important for efficient segmentation transformers, as it enables information transfer across different network depths. 
Similar ideas have shown success in the past, for example, hierarchical dense prediction transformers utilize cross-stage feature maps~\cite{xie2021segformer,wang2021pvt}.

We revisit Knowledge Review~\cite{chen2021knowledge_review}, which studies the connections across different levels of the teacher and the student CNN models.
However, the Attention Based Fusion (ABF), the original feature map fusion module in Knowledge Review, dynamically aggregates feature maps via a \textit{spatial} attention map, yet, overlooks the channel-wise information.
\rev{We argue that the teacher models typically have more blocks for feature extraction and more channels for retaining information, but they do not differ in spatial resolution compared to student models. Therefore, we aim to bridge the channel-wise differences between the feature maps of student and teacher models while incorporating spatial information through the design of a complementary \textit{channel} selection module.}
Our design is motivated by the following two aspects:
1) The channel-wise cues of the feature maps provide informative priors for feature extractors. Numerous previous works~\cite{li2019selective_kernel_networks,hu2018squeeze_excitation_network}, which focus on feature map fusion, seek to fuse spatial and channel information via channel attention.
Furthermore, Shu~\textit{et al.}~\cite{shu2021channel_distillation} indicate that minimizing the channel-wise Kullback-Leibler (KL) divergence of the feature- and prediction maps is more efficient than using its spatial counterpart.
2) As patch embedding distillation tracks dependencies among distant spatial positions via sequence-wise learning, feature map distillation should engage more spatially holistic knowledge in channels. It is therefore important to highlight channels from certain stages via reweighting and to account for cross-stage channel interdependencies.
To achieve this, we propose the \emph{Cross Selective Fusion (CSF)} module to adaptively fuse the information across stages via channel attention. 

\begin{figure}
    \centering
    \includegraphics[scale=0.7]{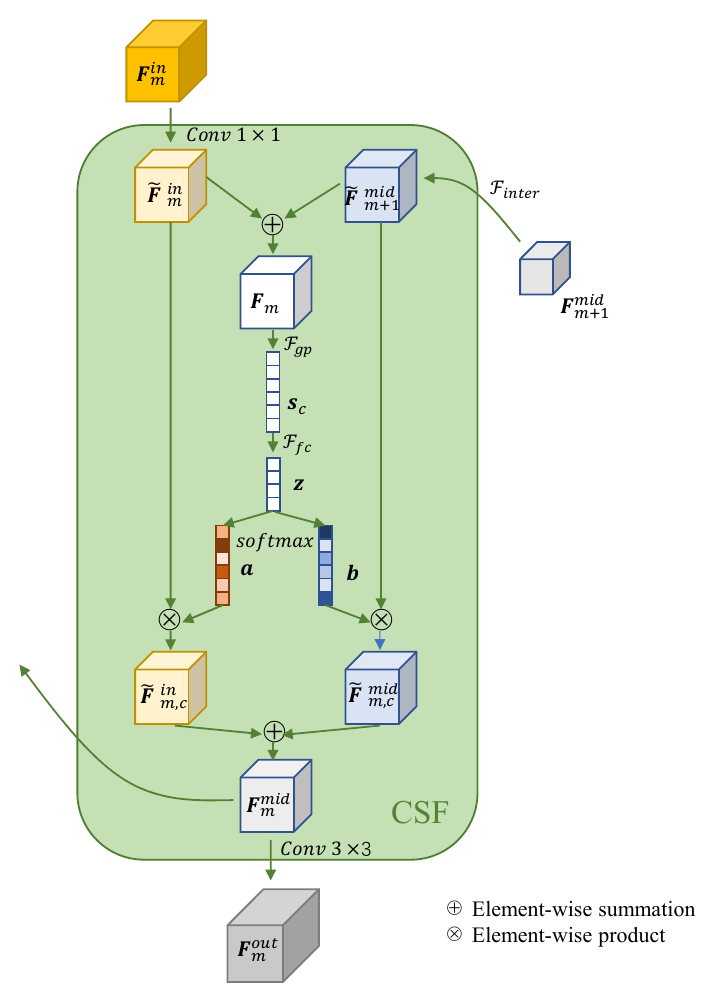}
    \caption{Architecture of the feature map fusion module, Cross Selective Fusion (CSF) in Fig.~\ref{fig:KD_structure}. The dimension-wise unified feature maps from the current student transformer block and the higher-level features are fused via channel attention.}
    \label{fig:csf}
\end{figure}
\noindent\textbf{Cross selective fusion.}
In this module, the feature maps from the current student stage and the next CSF are first resized to a uniform dimensionality and then fused via channel-wise attention.
As previously demonstrated in~\cite{li2019selective_kernel_networks}, feature extractions with different kernel sizes or modalities can be fused in an adaptive way.
Similarly, we propose to fuse cross-stage feature maps for a relation-based distillation.
As shown in Fig.~\ref{fig:csf}, we first conduct a $1{\times}1$ convolution to implement the channel-wise transformation of the input feature map $\mathbf{F}_{in}^m$ and conduct interpolation to implement the spatial transformation of the feature map $\mathbf{F}_{mid}^{m+1}$ from CSF of the next stage. Their height and width are the same as the feature map of the $m$-th stage $\mathbf{F}_{in}^m$. The resulting ${\widetilde{\mathbf{F}}^{in}}_{m}$ and ${\widetilde{\mathbf{F}}^{mid}}_{m+1}$ are element-wise summed, as depicted in Eq.~(\ref{eq:summation}):
\begin{equation}\label{eq:summation}
\mathbf{F}_m = {\widetilde{\mathbf{F}}^{in}}_{m} + {\widetilde{\mathbf{F}}^{mid}}_{m+1}.
\end{equation}
The global information is then embedded via global average pooling, where $\bm{s}_c$ denotes the channel-wise statistics, as shown in Eq.~(\ref{eq:global_average_pooling}):
\begin{equation}\label{eq:global_average_pooling}
{\mathbf{s}}_{c} = \mathcal{F}_{gp}(\mathbf{\bm{F}}_m) = \frac{1}{H \times W}\sum_{i=1}^{H}\sum_{j=1}^{W} \mathbf{F}_m(i,j).
\end{equation}
Then, a compact feature $\bm{z} \in \mathbb{R}^{d \times 1}$ in Eq.~(\ref{eq:z}) is generated for precise and adaptive selections:
\begin{equation}\label{eq:z}
\mathbf{z} = \mathcal{F}_{fc}(\mathbf{s}_c) = \delta(\mathcal{B}(\phi( \mathbf{s}_c))), \ \ \ d=\max(C/r,L),
\end{equation}
where $\delta (\cdot)$ denotes the ReLU function, $\mathcal{B}(\cdot)$ denotes Batch Normalization, and $\phi (\cdot)$ denotes $1{\times}1$ convolution. $d$ is the dimension of the vector $\bm{z}$, which is controlled by a reduction ratio $r$. $L$ is chosen to be the minimum value of $d$.

Soft attention across the channels is then  used to adaptively select information from different branches:
\begin{equation}
{a}_c = \frac{e^{\mathbf{A}_c \mathbf{z}}}{e^{\mathbf{A}_c \mathbf{z}} + e^{\mathbf{B}_c \mathbf{z}}},\ \  {b}_c = \frac{e^{\mathbf{B}_c \mathbf{z}}}{e^{\mathbf{A}_c \mathbf{z}} + e^{\mathbf{B}_c \mathbf{z}}}, 
\end{equation}
\begin{equation}
\mathbf{F}_{m,c}^{mid} = {a}_c \cdot {\widetilde{\mathbf{F}}^{in}}_{m} + {b}_c \cdot {\widetilde{\mathbf{F}}^{mid}}_{m+1}, \ \ \ {a}_c + {b}_c = 1,
\end{equation}
where $\mathbf{A},\mathbf{B} \in {\mathbb{R}}^{C \times d}$ are two learnable matrices, and $\mathbf{A}_c, \mathbf{B}_c \in {\mathbb{R}}^{1 \times d}$ are their  $c$-th elements. $a_c$, $b_c$, and $\mathbf{F}_{mid,c}^{m}$ are the $c$-th elements of $\bm{a}$, $\bm{b}$, and $\mathbf{F}_{mid}^{m}$. $\bm{a}$ and $\bm{b}$ denote the soft attention vector for ${\widetilde{\mathbf{F}}^{in}}_{m}$ and ${\widetilde{\mathbf{F}}^{mid}}_{m+1}$, respectively.
Furthermore, we perform a $3{\times}3$ convolution to transform $\mathbf{F}_{mid}^{m}$ to $\mathbf{F}_{out}^{m}$, ensuring it has the same dimension as the teacher's feature map.

For training, we compute the \rev{hierarchical context loss (HCL)}~\cite{chen2021knowledge_review} between the outputs of the CSF and the teacher transformer block at each stage, as shown in Fig.~\ref{fig:KD_structure}.
\rev{The feature map obtained from CSF at each stage, with shape $\left[n,c,h,w\right]$, is divided into $4$-level context information via adaptive average pooling. The height of the resulting multi-level feature maps at one stage is $\left[h,4,2,1\right]$, indicating that the shapes of the four abstract feature maps are $\left[n,c,h,w\right]$, $\left[n,c,4,4\right]$, $\left[n,c,2,2\right]$, and $\left[n,c,1,1\right]$.
At each stage, the $L_2$ distances are utilized to distill knowledge between the context levels. 
The final HCL loss between the feature maps at the $m$-th stage is computed as:
\begin{equation}\label{eq:hcl}
L_{fm}^m = {\frac{1}{1+{\sum_l}{\frac{1}{2^l}}}}{\sum_{l=0}^3}{\frac{1}{2^l}} MSE(\mathbf{F}_{m,l}^{out},\mathbf{F}_{m,l}^T),
\end{equation}
where $l$ refers to the number of levels at one stage.
The purpose of applying the HCL loss after every stage is to encourage knowledge distillation at different levels of abstraction.} 

\subsection{Configuration} 
\label{sec:configuration}
The core of our TransKD framework involves two knowledge distillation strategies: patch embedding distillation (in Sec.~\ref{sec:patch_embedding_distillaton}) and feature map distillation (in Sec.~\ref{sec:feature_map_distillaton}). 
To realize this design, we introduce four modules: \textit{PEA}, \textit{GL-Mixer}, and \textit{EA} modules guide the patch embedding distillation, while the \textit{CSF} feature map fusion module enhances the cross-stage relation-based feature map distillation.

Overall, CSF and PEA can be viewed as the fundamental components of TransKD.
The GL-Mixer operates on the \textit{deepest} representative embedding, (\textit{i.e.}, the final fourth stage), producing highly heterogeneous patterns~\cite{xie2021segformer}. It reinforces the patch embedding distillation at the final stage by harvesting global and local context information through two branches.
The EA modules are placed \textit{at each stage} and serve as an embedding method to provide supplementary distillation streams by using an identical embedding method to the teacher's and the student's, with the teacher's number of channels.
It should be taken into account that, as different transformer architectures have different embedding methods, the framework leveraging EA does not yield full plug-and-play flexibility. 

We implement and examine different variants of our proposed framework.
Note, that we cannot isolate and verify the efficacy of each module when all the modules are combined together, as an integration of these components means many constraints and could lead to overfitting.
We, therefore, focus on three configurations of feature map distillation and patch embedding distillation provided in Tab.~\ref{tabel:configuration}, which we refer to as \textit{TransKD-Base}, \textit{TransKD-GL}, and \textit{TransKD-EA}. The structure of these three variants is provided in Fig.~\ref{fig:KD_structure}.

\begin{table}[h]
\setlength{\tabcolsep}{4.5pt}
\caption{Configuration of TransKD variants. \rev{\textbf{FM:} Feature Map. \textbf{PE:} Patch Embedding.}}
\begin{center}
\input{tables/configuration}
\end{center}
\label{tabel:configuration}
\end{table}

\section{Experiment Datasets and Setups}
\input{Tex_content/experiment_datasets_and_setups}

\section{Experiment Results and Analyses}

\subsection{Quantitative results}
\label{sec:quantitative_results}

\noindent\textbf{Results on the Cityscapes dataset.}
In Tab.~\ref{tabel:ComparationKD}, we compare the performance of our TransKD method without pre-training the student model on ImageNet, to a multitude of previously published knowledge distillation approaches, on the Cityscapes validation set.
We first validate the usefulness of our framework against a baseline trained without knowledge distillation and see that a student SegFormer-B0 model optimized with our TransKD-EA distillation framework achieves a $+13.12\%$ improvement ($55.86\%$ \textit{vs.} $68.98\%$) in mIoU.
TransKD-EA also outperforms other distillation frameworks by a significant margin, for example, $+5.58\%$ compared to the feature-map-only distillation approach Knowledge Review (KR)~\cite{chen2021knowledge_review}, revealing the benefits of unifying patch embedding and feature map distillation.
Furthermore, the student model distilled by using TransKD-EA delivers results comparable to the student model with additional lengthy pre-training.
Although our TransKDs seem to be complex, they are not that heavy compared to SKDs~\cite{liu2019structured_knowledge_distillation} and achieve plug-and-play knowledge distillation. 
Still, the lightweight variant of our framework TransKD-Base is alone sufficient to conduct the distillation, yielding a surprising $+5.18\%$ gain compared to KR~\cite{chen2021knowledge_review} while just adding $0.21M$ parameters for patch embedding distillation. 

To explore the two knowledge sources, feature maps and patch embeddings, equally, we perform Knowledge Review on the two positions, respectively. The results show that distilling the knowledge from patch embeddings has a decent gain compared to distilling from feature maps ($63.40\%$ \textit{vs.} $62.41\%$). The knowledge from patch embeddings can complement but not replace those from feature maps. Besides, it turns out that the review mechanism may not be perfectly suitable for positional information.
\begin{table}[t!]
\setlength{\tabcolsep}{5.0pt}
\caption{Comparison with knowledge distillation methods on the Cityscapes dataset~\cite{cordts2016cityscapes}. * denotes at the position of patch embeddings.}
\begin{center}
\input{tables/comparison_with_KD_methods}
\end{center}
\label{tabel:ComparationKD}
\end{table}
\noindent\textbf{Comparison against previous distillation methods.}
Compared to the traditional response-based~\cite{shu2021channel_distillation,hinton2015distilling_knowledge}, feature-based~\cite{liu2019structured_knowledge_distillation,shu2021channel_distillation}, and relation-based~\cite{chen2021knowledge_review} knowledge distillation, the combination of traditional distillation and patch embedding distillation greatly improves the effectiveness of transformer-to-transformer transfer.
Besides, compared to the teacher model which is based on the SegFormer-B2, the student model with SegFormer-B0 backbone reduces ${>}85\%$ GFLOPs.
Moreover, TransKD-EA rivals the notoriously time-consuming pre-training method.
Note that the number of parameters and GFLOPs listed in Tab.~\ref{tabel:ComparationKD} include those of the model itself and those of the distillation framework.

However, at test-time, the computational complexity of the enhanced student model is not affected by the distillation method and is identical to the original compact student model ($3.72M$ parameters and $13.67$ GFLOPs), therefore keeping the high efficiency needed in real-world applications.

\begin{table*}[t!]
   \caption{Per-class IoU scores of our proposed knowledge distillation frameworks compared with the feature-map-only baseline distillation method Knowledge Review~\cite{chen2021knowledge_review} on the validation set of Cityscapes~\cite{cordts2016cityscapes}.
   }
	\renewcommand\arraystretch{1.1}
	%\footnotesize
	\centering
    \input{tables/cityscapes_class}
	\label{tab:class-iou-table}
\end{table*}

\begin{figure}[!t]
\centering
\subfloat[Per-class IoU results on Cityscapes\label{subfig-1:class_iou_cityscapes}]{%
  \includegraphics[width=0.45\textwidth]{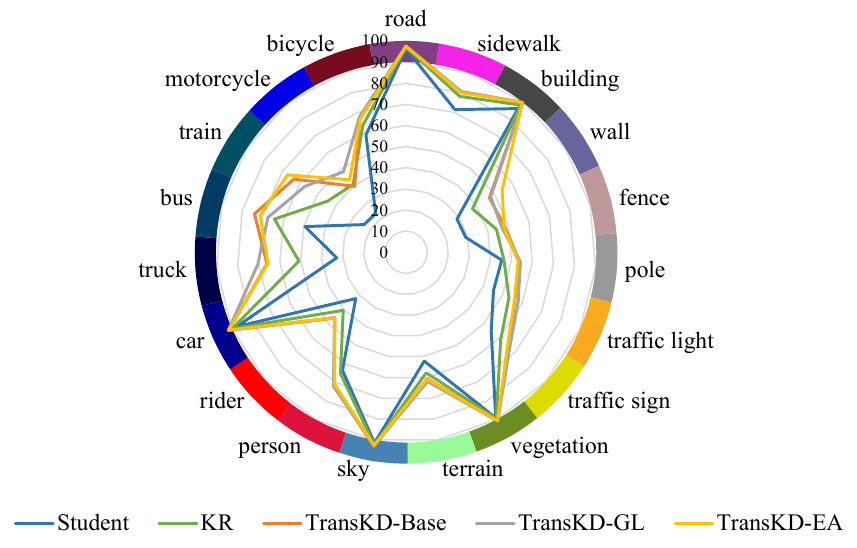}
}
\hfill

\subfloat[Per-class IoU results on ACDC\label{subfig-2:class_iou_acdc}]{%
  \includegraphics[width=0.45\textwidth]{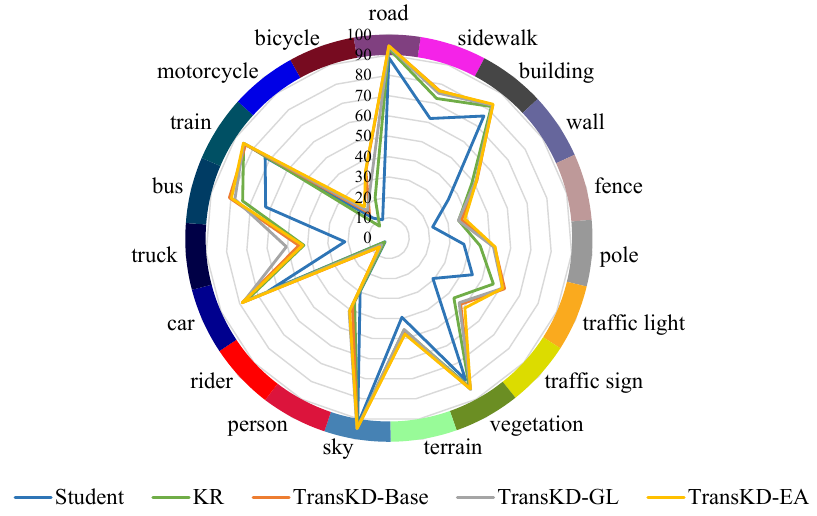}
}
\caption{Semantic segmentation IoUs of our TransKDs compared with non-pretrained student and Knowledge Review~\cite{chen2021knowledge_review} on Cityscapes and ACDC datasets.}
\label{fig:class_iou}
\vskip -4ex
\end{figure}

\noindent\textbf{Per-class accuracy analyses on Cityscapes.}
Next, we investigate the performance of our framework for the individual categories in Tab.~\ref{tab:class-iou-table}, comparing the TransKD variants to the feature-map-only distillation baseline Knowledge Review (KR)~\cite{chen2021knowledge_review}.
TransKD-GL is better at recognizing small infrastructures, such as \emph{traffic light}, \emph{traffic sign}, and \emph{pole}, while the TransKD-EA has better performance on the categories like \emph{wall} and \emph{train}.
Both variants surpass the KR framework for all classes with a remarkable gap, especially for \emph{wall}, \emph{pole}, \emph{person}, \emph{rider}, \emph{truck}, and \emph{train}.
In Fig.~\ref{fig:class_iou}(a), we further showcase the per-class accuracy of the original student model, the KR framework, and our TransKD approaches.
The benefits of TransKDs are especially large for the challenging less-frequent vehicle categories such as \emph{bus}, \emph{train}, and \emph{truck}.

\noindent\textbf{Effectiveness for different architectures.}
We then verify whether the effectiveness of our method is consistent for different architectures.
With this intent, we compare the TransKD-Base method against the Knowledge Review (KR) framework~\cite{chen2021knowledge_review} for different segmentation backbones on the Cityscapes dataset.
The results in Tab.~\ref{tab:universe} indicate that our approach is not strictly tied to a concrete model, but is consistently beneficial for different segmentation backbones.
Specifically, our TransKD-Base method achieves $+5.18\%$, $+2.18\%$, and $+2.71\%$ improvements over the feature-map-only KR method, while using the SegFormer-B0, the PVTv2-B0~\cite{wang2022pvt_v2}, and the Lite Vision Transformer (LVT)~\cite{yang2022lite_vision_transformer} model as the student, respectively.
For the PVTv2-B0, we use PVTv2-B2~\cite{wang2022pvt_v2} as the teacher model.
In the setting of using the SegFormer-B2 as the teacher and the LVT as the student (which have rather dissimilar structures), our proposed TransKD framework also clearly improves the segmentation quality.
When comparing to the original student segmenter without any distillation, the gains are more evident with $12.72\%$, $10.34\%$, and $11.05\%$ in mIoU. 
These results show that our TransKD framework can unleash the potential of the efficient transformer models, yielding sufficient improvements of the compact models.

\rev{
Furthermore, we assess the efficiency of various transformer architectures during both training and inference phases. During training, the complexity of teacher and student transformers embedded within knowledge distillation frameworks is measured in terms of the number of parameters and GFLOPs. Conversely, during the inference phase, the frame rate per second (FPS) is evaluated using pure transformers not equipped with knowledge distillation frameworks on the identical GPU, as the frameworks are primarily employed for knowledge transfer during training only.
The student transformers utilized in this analysis are significantly smaller in size, ranging from $7.35$ to $3.86$ times smaller compared to the teacher models. Moreover, incorporating the distillation framework incurs only a minimal increase in the number of parameters (approximately $0.84$ million parameters) and computational cost (an additional $2.97$ GFLOPs), irrespective of the model type. Notably, the inference speed of the student models is enhanced by a factor of $2$ to $3.5$ times relative to that of the teacher models. Overall, student transformers offer faster inference with minimal additional cost and reduced sizes.}

\begin{table}[t!]
    \centering
    \caption{Accuracy analysis on various transformers.}
    \input{tables/various_architectures}
    \label{tab:universe}
\end{table}
\noindent\textbf{Results on the ACDC dataset.}
Tab.~\ref{table_acdc} summarizes the distillation outcome of  CD~\cite{shu2021channel_distillation}, KR~\cite{chen2021knowledge_review}, and our TransKD approaches on the ACDC validation dataset.
Apart from the overall mIoU scores, we examine the models on four different adverse conditions captured in the ACDC dataset, namely \emph{fog}, \emph{night}, \emph{rain}, and \emph{snow}.
Our TransKD-EA method obtains the best performance with $59.09\%$ of mIoU on the All-ACDC benchmark, and the top scores of $66.13\%$, $44.99\%$, and $57.24\%$ of mIoU for the \emph{fog}, \emph{night}, and \emph{rain} subsets respectively.
Interestingly, another variant of our framework, TransKD-GL, yields the best outcome of $60.61\%$ on the \emph{snow} subset of the ACDC dataset.
We hypothesize that for snowy scenes, local cues are crucial for segmentation and TransKD-GL excels, due to the extraction of detailed information from the representative high-level embedding.
Compared to Knowledge Review, all of our three TransKD variants have respective $+3.66\%$, $+3.23\%$, and $+4.19\%$ performance gains.
The improvements in these diverse weather and nighttime subsets confirm the effectiveness of TransKD for different scene appearances.

\begin{table}[t!]
%\scriptsize 
\setlength{\tabcolsep}{4.5pt}
\caption{Accuracy analysis on the ACDC dataset~\cite{sakaridis2021acdc} in different adverse conditions.}
\label{table_acdc}
\begin{center}
\input{tables/acdc}
\vskip -4ex
\end{center}
\end{table}

\noindent\textbf{Per-class accuracy analyses on ACDC.}
We also provide the per-class segmentation results of different frameworks on the ACDC dataset in Fig.~\ref{fig:class_iou}(b).
Compared to the original student and the KR-boosted student, models optimized with our TransKD approach have evident gains in most classes.
In particular, TransKD has substantial improvements on certain categories that take less space (and are thus harder to segment) but are very important for autonomous driving, such as \emph{pole}, \emph{traffic light}, and \emph{traffic sign}.

\noindent\textbf{Results on the NYUv2 dataset.}
Our next area of investigation is segmentation results for the \textit{indoor} scenes covered by the NYUv2 dataset~\cite{silberman2012nyu_dataset}.
The original student model with SegFormer-B0 achieves a relatively low performance of $18.19\%$ in mIoU on NYUv2.
While the CD and KR frameworks only bring limited improvements, our TransKD-GL method reaches the best mIoU score of $24.20\%$.
Compared to KR, our TransKD-Base, -GL, and -EA distillation frameworks reach $+0.81\%$, $+1.40\%$, and $+1.09\%$ mIoU gains, respectively.
This indicates that our TransKD methods are also beneficial for indoor scene understanding.

\begin{table}[h]
%\scriptsize %\setlength{\tabcolsep}{2.0pt}
\setlength{\tabcolsep}{5.0pt}
\caption{Accuracy analysis on the NYUv2 dataset~\cite{silberman2012nyu_dataset}.}
\label{table_nyu}
\begin{center}
\input{tables/nyu}
\end{center}
\end{table}

\noindent\textbf{Results on the Pascal VOC2012 dataset.}
As both Pascal VOC2012 and ImageNet are object-centric datasets with similar layouts, training with our TransKD is assumed to be unable to obtain the common pattern from the pre-training process on ImageNet. However, knowledge distillation through our TransKD compensates for $81.1\%$ of the gap between the non-pretrained and pretrained models successfully, which means that our TransKD-EA achieves a $+19.3\%$ gain over the student model. All three variations, TransKD-Base, TransKD-GL, and TransKD-EA attain salient improvements of $+4.17\%$, $+4.49\%$, and $+4.68\%$ in mIoU over the baseline solution Knowledge Review~\cite{chen2021knowledge_review}.
\begin{table}[t]
    \centering
    \caption{Accuracy analysis on the Pascal VOC2012 dataset~\cite{Everingham2010voc}. 
    }
    \label{tab:voc2012}
    \begin{center}
        \input{tables/voc2012}
    \end{center}
\end{table}
\subsection{Ablation Studies and Comparisons}
\noindent\textbf{Effectiveness of feature map fusion module and loss function.}
The feature map distillation component of TransKD (Sec. \ref{sec:feature_map_distillaton}) covers the feature map fusion modules and the loss functions.
For robust feature map distillation, we have introduced Cross Selective Fusion (CSF), to substitute the original feature fusion module ABF in Knowledge Review~\cite{chen2021knowledge_review}. 
CSF is based on a channel attention mechanism for transformer feature map distillation.
To verify the effectiveness of CSF, we compare CSF and ABF when using the HCL loss function on Cityscapes~\cite{cordts2016cityscapes} and ACDC~\cite{sakaridis2021acdc}  in Tab.~\ref{table:ffm}.
It is evident, that CSF clearly outperforms ABF with accuracy gaps of $2.54\%$ and $0.45\%$ on the two datasets.
This finding showcases the usefulness of CSF and the importance of channel-wise dependencies for transformer-based knowledge distillation.

\begin{table}[h]
%\scriptsize
\caption{Accuracy analysis on the feature map fusion module.}
\label{table:ffm}
\begin{center}
\input{tables/fusion_module}
\end{center}
\vskip -2ex
\end{table}

We further aim to isolate the role of the feature map loss function in a separate set of experiments, with the results summarized in Tab.~\ref{table:lf}.
The Kullback-Leibler (KL) divergence loss is widely used in knowledge distillation and enables the student to imitate the teacher's distributions, \textit{i.e.} model predictions~\cite{shu2021channel_distillation,liu2019structured_knowledge_distillation,hinton2015distilling_knowledge} or feature maps~\cite{shu2021channel_distillation}.
Thus, we choose the channel-wise and spatial KL divergence loss functions as control groups for confirming the effectiveness of HCL (the hierarchical version of the MSE loss).
The results demonstrate that HCL outperforms both KL divergence losses by ${>}1.5\%$ when distilling the intermediate feature maps.

\begin{table}[t]
%\scriptsize
\caption{Accuracy analysis on the feature map distillation loss function.}
\label{table:lf}
\begin{center}
\input{tables/loss_function}
\end{center}
\vskip -4ex
\end{table}

\noindent\textbf{Effectiveness of patch embedding distillation.}
In this paper, patch embedding distillation (Sec. \ref{sec:patch_embedding_distillaton}) is the first introduced in the field of semantic segmentation.
Tab.~\ref{table:PEAbyStage} summarizes our experiments targeting the effects of patch embedding distillation for different stages.
When distilling one stage of patch embedding, the patch embedding loss is directly added to the feature map loss, \textit{i.e.}, ${\alpha_m}=1$.
Patch embedding distillation at any stage is effective, as it clearly boosts the performance compared to its counterparts without knowledge exchange at the patch embedding level.
Generally, the deeper the stage, the more channels of the patch embedding are used and the better the distillation effect the framework can provide.
Note that the original feature-map-only Knowledge Review does not use any patch embedding distillation and reaches a much lower outcome of $63.40\%$. 
Models yield their best performances with all-stage patch embedding distillations in place, \textit{e.g.}, $67.89\%$ in mIoU for Knowledge Review (equipped with our PEA modules) and $68.58\%$ for TransKD. 

Additionally, here we use CSF+HCL to represent our feature map distillation part in TransKD-Base and compare it with Knowledge Review combined with PEA.
Our CSF+HCL improves the outcome by $1.2\%{\sim}2.5\%$ over Knowledge Review with different PEA configurations (both when it is not combined or combined with  PEA at different stages), verifying the benefits of the proposed feature map fusion module CSF.
When the feature map distillation frameworks leverage PEA at all stages with weights of $\bm{\alpha}=[0.1,0.1,0.5,1]$, CSF+HCL reaches the best mIoU of $68.58\%$ still surpassing the Knowledge Review variant enhanced with our patch embedding distillation blocks by $0.69\%$.

\rev{Furthermore, we summarize the performances of TransKDs across various datasets in Tab.~\ref{tab:op_modules} to analyze the effectiveness of the optimization modules, GL-Mixer and EA. We observe that the optimization modules consistently improve the performance of the basic version of TransKD by over $+0.5\%$. However, while TransKD enhanced by GL-Mixer underperforms the basic version on the ACDC dataset, TransKD-GL achieves state-of-the-art results in the \textit{snow} condition, as demonstrated in Tab.~\ref{table_acdc}.}

\begin{table}[t!]
%\scriptsize
\setlength{\tabcolsep}{3.5pt}
\caption{Accuracy analysis on integration of PEA by stages.}
\label{table:PEAbyStage}
\begin{center}
\input{tables/PEA_by_stage}
\end{center}
\vskip -4ex
\end{table}

\begin{table}[t]
\centering
\rev{
\caption{Accuracy analysis of optimization modules for patch embedding distillation.}
\input{tables/comparison_op_versions}
\label{tab:op_modules}
}
\vskip -4ex
\end{table}

\noindent\textbf{Influence of input size and pre-training weights.}
According to the original work~\cite{xie2021segformer}, using large input images significantly improves the segmentation results of SegFormer. Besides, ImageNet pre-training is a ubiquitous way for transformer-based methods to compensate for the lack of inductive bias and speed up convergence.
In Tab.~\ref{tab:fullsize}, we examine the influence of the input size and pre-trained weights by comparing the results of Knowledge Review~\cite{chen2021knowledge_review} and TransKD-Base for different configurations.

We first look at the networks \textit{without} weight initialization via pre-training, (\textit{i.e.}, without ``+ImN'' in Tab.~\ref{tab:fullsize}).
In this case, the large image scale facilitates the efficacy of the overall semantic segmentation model while compromising the gain of TransKD relative to Knowledge Review and the student.
At the image size of $512{\times}1024$, the TransKD-distilled student raises the accuracy by $12.72\%$ compared to the original non-pretrained student.
With the full size of $1024{\times}2048$, our TransKD-Base attains the performance of $71.59\%$ with a less, yet still significant accuracy increase of $9.73\%$ in mIoU.
\begin{table}[t!]
    \centering
    \caption{Accuracy analysis on different sizes of input images. 
    }
    \input{tables/fullsize}
    \label{tab:fullsize}
\vskip -4ex
\end{table}

\begin{table}[t!]
\scriptsize
    \centering
    \caption{Performance comparison with state-of-the-art distillation methods over different student segmentation networks on the Cityscapes validation set~\cite{cordts2016cityscapes}. * denotes that the efficient segmentation model uses pre-trained weights. ``-'' denotes that the corresponding information is missing in the respective paper.}
    \input{tables/comparison_other_result}
    \label{tab:comparison_other_result}
\end{table}

As expected, a larger input size leads to better results for both, the pre-trained and the non-pretrained networks.
We believe, that since pre-training on a large dataset is supposed to complement the inductive bias, 
and the proposed patch embedding distillation follows the same purpose, 
patch embedding distillation can compensate for the gap between the non-pretrained and the pretrained transformers sufficiently, but does not show a salient improvement over the pre-trained transformer. Thus,
TransKD-Base outperforms Knowledge Review by a smaller margin when the students are pre-trained.
Unsurprisingly, the best segmentation quality is achieved when fine-tuning the student pre-trained on ImageNet with our TransKD-Base knowledge distillation approach, yielding $75.74\%$ in mIoU.

\noindent\textbf{Comparison to state-of-the-art segmenters.}
In Tab.~\ref{tab:comparison_other_result}, we further compare TransKD-Base against state-of-the-art semantic segmentation networks and previous knowledge distillation methods.
While heavier models like CCNet~\cite{huang2019ccnet} and SETR~\cite{zheng2021setr}  reach high mIoU scores, they are computationally intensive, which often disqualifies their deployment in mobile and real-time systems.
Accurate segmenters like HANet~\cite{choi2020cars_hanet} and SegFormer-B5~\cite{xie2021segformer} require ${>}1000G$ FLOPs to yield precise high-resolution predictions.
We further consider more efficient semantic segmenters. 
Compact models, such as CGNet~\cite{wu2020cgnet} and Fast-SCNN~\cite{poudel2019fastscnn}, are very lightweight but come at a high accuracy cost, and the segmentation quality declines.
Some models with moderate computation complexity like ERFNet~\cite{romera2018erfnet} and BiSeNet~\cite{yu2018bisenet} can attain higher accuracy but also comprise more model parameters.

\begin{figure*}[!t]
\centering
\subfloat[Non-pretrained Networks\label{comparison_other_subfig-1:nonpretrained}]{%
  \includegraphics[width=0.35\textwidth]
  {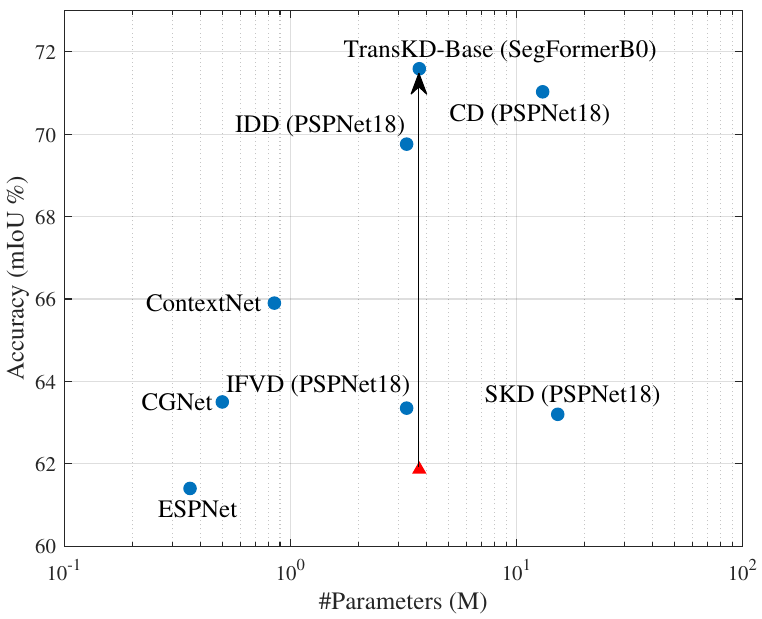}
}
\hfil
\subfloat[Pre-trained Networks\label{comparison_other_subfi-2:pretrained}]{%
  \includegraphics[width=0.35\textwidth]
  {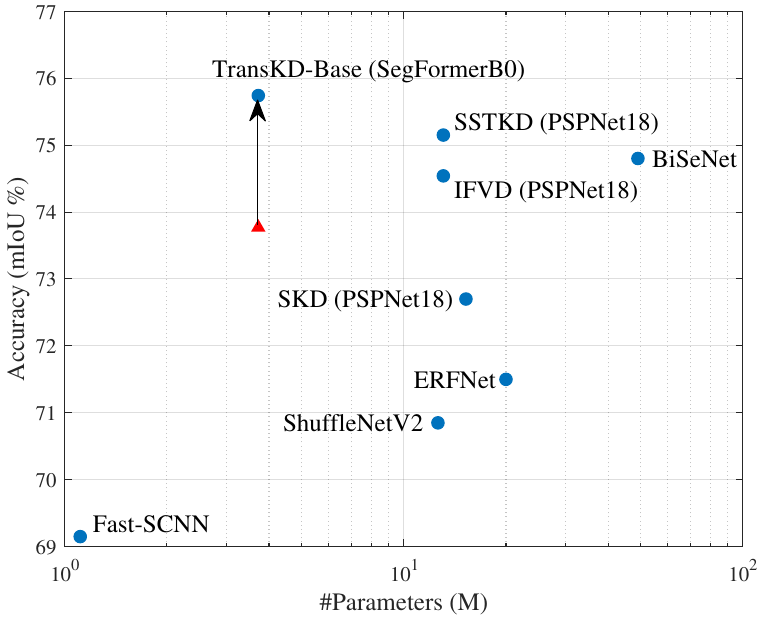}
}
\caption{Comparison between TransKD and state-of-the-art efficient semantic segmentation networks and knowledge distillation methods, in terms of (a) not using pre-trained weights and (b) using pre-trained weights. TransKD greatly boosts the mIoU of the efficient model.}
\label{fig:other_results}
\vskip -2ex
\end{figure*}

Our previous experiments in Sec.~\ref{sec:quantitative_results} examined different knowledge distillation methods with the same student backbone.
In this section, we conduct more comparisons to previously published methods with numbers from their respective papers for the enhanced student after distillation (Tab.~\ref{tab:comparison_other_result}).
Most of the previous works have used PSPNet18~\cite{zhao2017pspnet} as the student model.
In the group of distillation methods and non-pretrained students, our TransKD-Base yields a top performance of $71.59\%$ in mIoU, which is even close to the results of some pre-trained students, \textit{e.g.}, MD~\cite{xie2018improving_md} ($71.90\%$).
In the group of distillation methods and pre-trained students, TransKD achieves the highest mIoU score despite being the most lightweight model with only $3.72M$ parameters and $54.70G$ FLOPs.
Fig.~\ref{fig:other_results} visualizes the comparison of representative efficient segmentation models and knowledge distillation methods. We clearly see that TransKD stands out in front of the past state-of-the-art models in both groups and strikes an excellent trade-off between efficiency and accuracy for semantic scene understanding.
\subsection{Qualitative Results}

\begin{figure}[h]
\includegraphics[width=0.48\textwidth]{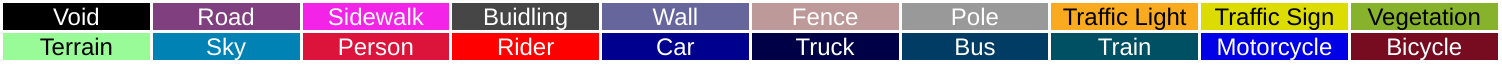}
\subfloat[Student w/o KD\label{tsne-subfig-1:w/oKD}]{%
  \includegraphics[width=0.155\textwidth]{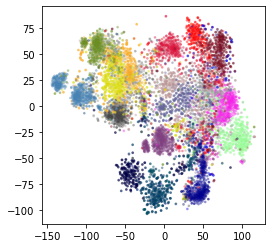}
}
\hfill
\subfloat[KR \label{tsne-subfig-2:withKR}]{%
  \includegraphics[width=0.155\textwidth]{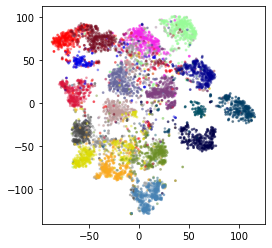}
}
\hfill
\subfloat[TransKD-Base\label{tsne-subfig-2:withTransKD}]{%
  \includegraphics[width=0.155\textwidth]{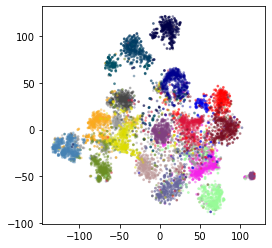}
}
\caption{t-SNE visualization of the feature maps at the last stage. A random sampling of five hundred feature nodes is taken in each category. The prediction maps are down-sampled to label the feature nodes semantically.}
\label{fig:TSNE}
\vskip -2ex
\end{figure}

\noindent{\textbf{Analyses of t-SNE visualizations.}}
We next consider the representation point of view and asses the discriminability of the learned embeddings in the latent space via t-SNE visualizations~\cite{van2008visualizing} given in Fig.~\ref{fig:TSNE}, where Fig.~\ref{fig:TSNE}(a) illustrates the original student embeddings, and Fig.~\ref{fig:TSNE}(b) and Fig.~\ref{fig:TSNE}(c) are the student representations learned with KR~\cite{chen2021knowledge_review} and our TransKD-Base, respectively.

Without any knowledge distillation, the decision boundaries for most of the categories are obscure and difficult to distinguish, which leads to harder decision-making for the classifier. 
Training with Knowledge Review helps to distinguish most of the categories, but  \emph{traffic sign} and \emph{traffic light}, which are crucial for autonomous vehicles, are easily confused. Finally, these issues are better addressed by utilizing our TransKD knowledge distillation.
Compared to Fig.~\ref{fig:TSNE}(a) and Fig.~\ref{fig:TSNE}(b), we observe clearer inter-category boundaries in Fig.~\ref{fig:TSNE}(c), which indicates the benefits of the proposed methods unifying patch embedding- and feature distillation as it enhances the discriminability in the latent space. 

\begin{figure*}[t!]
% \footnotesize
\small
% \scriptsize
\setlength\tabcolsep{1pt}
\centering

\begin{tabular}{c c c c c c}
\rotatebox[origin=c]{90}{Image} &
\raisebox{-0.5\height}{\includegraphics[width=0.193\textwidth]{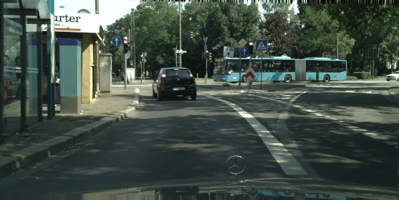}} &
\raisebox{-0.5\height}{\includegraphics[width=0.172\textwidth]{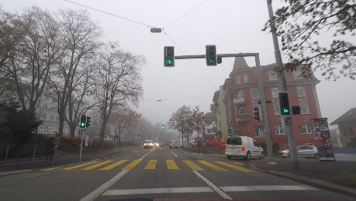}} &
\raisebox{-0.5\height}{\includegraphics[width=0.172\textwidth]{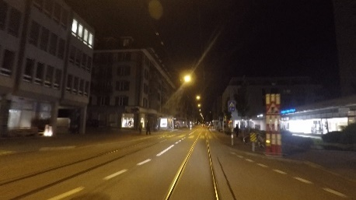}} &
\raisebox{-0.5\height}{\includegraphics[width=0.172\textwidth]{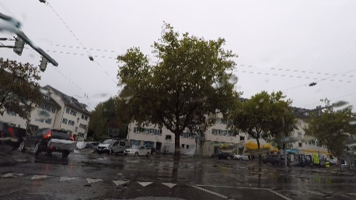}} &
\raisebox{-0.5\height}{\includegraphics[width=0.172\textwidth]{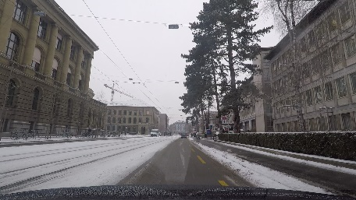}}\\
\rotatebox[origin=c]{90}{\makecell{Ground\\Truth}} &
\raisebox{-0.5\height}{\includegraphics[width=0.193\textwidth]{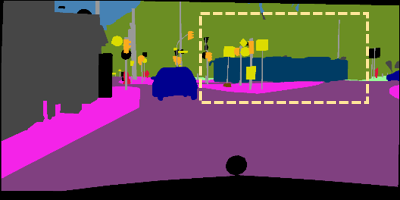}} &
\raisebox{-0.5\height}{\includegraphics[width=0.172\textwidth]{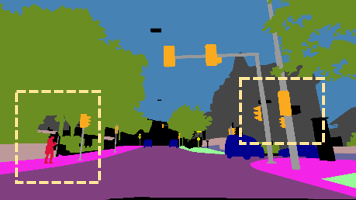}} &
\raisebox{-0.5\height}{\includegraphics[width=0.172\textwidth]{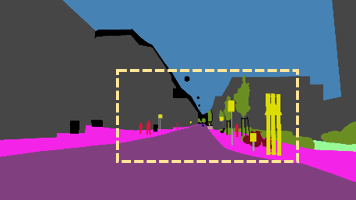}} &
\raisebox{-0.5\height}{\includegraphics[width=0.172\textwidth]{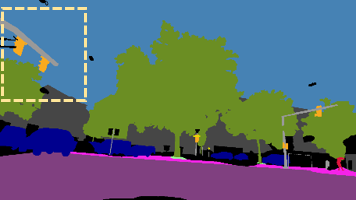}} &
\raisebox{-0.5\height}{\includegraphics[width=0.172\textwidth]{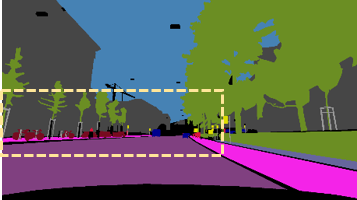}}\\
\rotatebox[origin=c]{90}{Student} &
\raisebox{-0.5\height}{\includegraphics[width=0.193\textwidth]{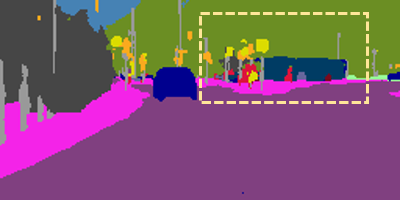}} &
\raisebox{-0.5\height}{\includegraphics[width=0.172\textwidth]{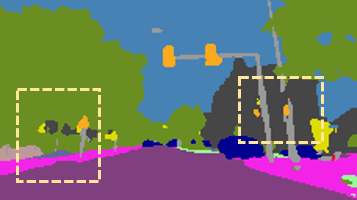}} &
\raisebox{-0.5\height}{\includegraphics[width=0.172\textwidth]{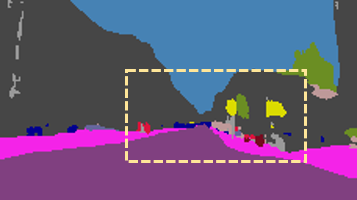}} &
\raisebox{-0.5\height}{\includegraphics[width=0.172\textwidth]{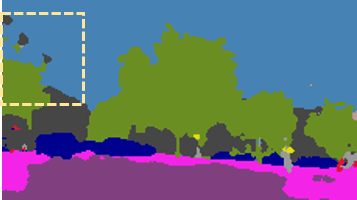}} &
\raisebox{-0.5\height}{\includegraphics[width=0.172\textwidth]{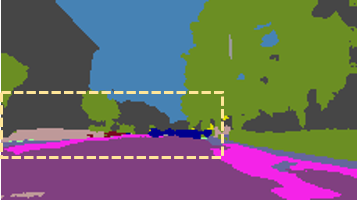}}\\
\rotatebox[origin=c]{90}{CD} & 
\raisebox{-0.5\height}{\includegraphics[width=0.193\textwidth]{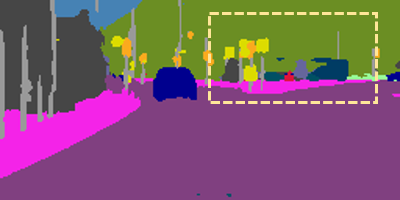}} &
\raisebox{-0.5\height}{\includegraphics[width=0.172\textwidth]{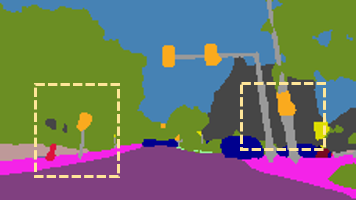}} &
\raisebox{-0.5\height}{\includegraphics[width=0.172\textwidth]{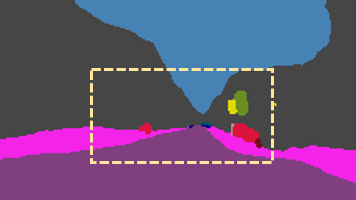}} &
\raisebox{-0.5\height}{\includegraphics[width=0.172\textwidth]{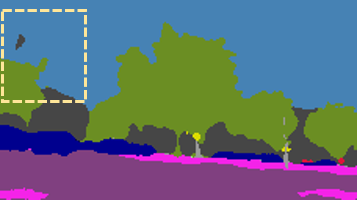}} &
\raisebox{-0.5\height}{\includegraphics[width=0.172\textwidth]{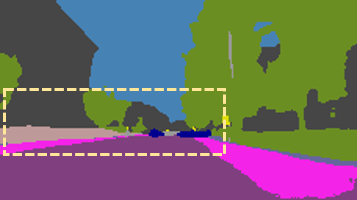}}\\
\rotatebox[origin=c]{90}{KR} & 
\raisebox{-0.5\height}{\includegraphics[width=0.193\textwidth]{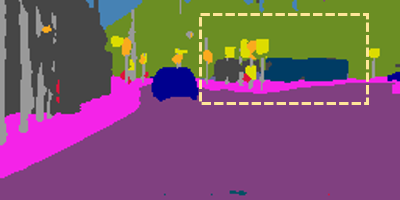}} &
\raisebox{-0.5\height}{\includegraphics[width=0.172\textwidth]{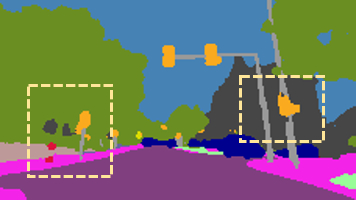}} &
\raisebox{-0.5\height}{\includegraphics[width=0.172\textwidth]{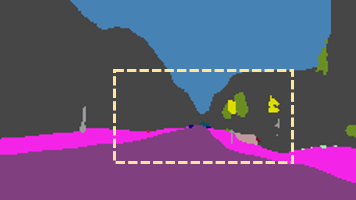}} &
\raisebox{-0.5\height}{\includegraphics[width=0.172\textwidth]{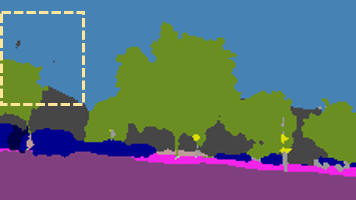}} &
\raisebox{-0.5\height}{\includegraphics[width=0.172\textwidth]{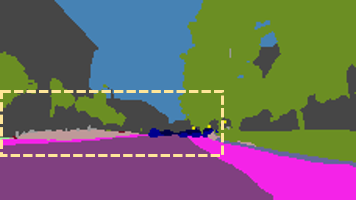}}\\
\rotatebox[origin=c]{90}{\makecell{TransKD}} &
\raisebox{-0.5\height}{\includegraphics[width=0.193\textwidth]{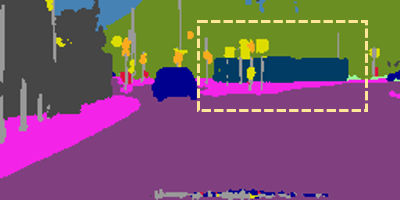}} &
\raisebox{-0.5\height}{\includegraphics[width=0.172\textwidth]{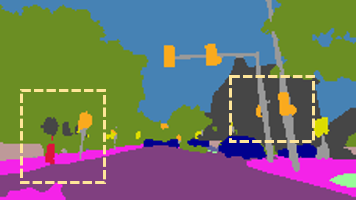}} &
\raisebox{-0.5\height}{\includegraphics[width=0.172\textwidth]{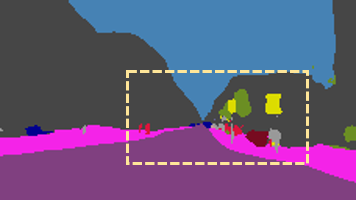}} &
\raisebox{-0.5\height}{\includegraphics[width=0.172\textwidth]{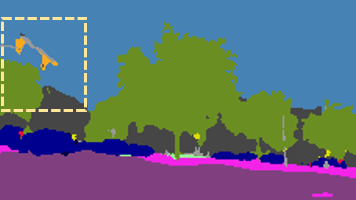}} &
\raisebox{-0.5\height}{\includegraphics[width=0.172\textwidth]{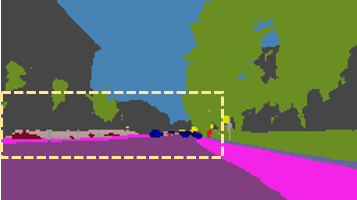}}\\
~&Normal&Fog&Night&Rain&Snow
\end{tabular}
\caption{Qualitative street scene semantic segmentation results under normal condition from the Cityscapes dataset~\cite{cordts2016cityscapes} and adverse conditions from the ACDC dataset~\cite{sakaridis2021acdc} including fog, night, rain, and snow conditions. The performance of our TransKD is compared to Knowledge Review (KR)~\cite{chen2021knowledge_review}. Here we choose the representative results of TransKD-EA.}
\label{fig:segmentation_map_visualization}
\end{figure*}

\noindent{\textbf{Analyses of scene segmentation effects.}}
Qualitative results of semantic segmentation in street scenarios under different kinds of weather conditions are depicted in Fig.~\ref{fig:segmentation_map_visualization}, where normal, foggy, nightly, rainy, and snowy weathers are considered from left to right in each row.
The samples come from Cityscapes (normal weather) and ACDC (adverse weather), respectively.
We compare TransKD 
against the feature-map-only Knowledge Review (KR)~\cite{chen2021knowledge_review}
and Channel-wise Distillation (CD)~\cite{shu2021channel_distillation}.
The challenging parts of each example are highlighted with the yellow dashed rectangles.
For the example featuring normal-weather yet a complex scene (the \emph{bus} occluded by \emph{traffic signs}), a large number of false predictions on small objects show up in the student's result without distillation, while similar unsatisfactory segmentation maps can be also found in CD and KR's outcomes, \textit{e.g.}, part of the \emph{bus} is wrongly classified as \emph{buildings}. Fortunately, these issues are better addressed by TransKD, leading to complete and much clearer segmentation and demonstrating its efficacy in dealing with details.

These findings are also confirmed in our experiments featuring different weather conditions of the ACDC benchmark.
In the foggy, nighttime, and rainy scenes, TransKD enables better detection of challenging small objects like safety-critical \emph{pedestrians}, as well as \emph{traffic lights} and \emph{traffic signs}.
In the \emph{snowy} scene, TransKD can also parse the \emph{vegetation} despite this class having similar textures as the background \emph{buildings}, thanks to the embedding distillation transferring long-range dependencies.
This again highlights the potential of linking patch embedding- and feature map distillation via our TransKD, which derives more informative knowledge than the feature-level-only
distillation approach KR.

\noindent
\rev{
\textbf{Limitation and future work.} Although TransKDs enable non-pretrained student models to perform comparably to their pretrained counterparts, there remains a minor performance gap ($0.77\%$ on Cityscapes) between them. In the future, we anticipate that the knowledge sources, such as patch embeddings and feature maps, which have proven effective in our work, can be better leveraged to close this performance gap through more refined knowledge distillation framework designs. Furthermore, TransKDs have been exclusively implemented on pyramid transformers, which are primarily used in semantic segmentation. We believe that our TransKD framework could also be adapted to isotropic vision transformers.
}
\section{Conclusion}
\input{Tex_content/conclusion}

\bibliographystyle{IEEEtran}
\bibliography{bib}

\input{Tex_content/supp}

\end{document}

%% file: Tex_content/abstract.tex
Semantic segmentation benchmarks in the realm of autonomous driving are dominated by large pre-trained transformers, yet their widespread adoption is impeded by substantial computational costs and prolonged training durations. To lift this constraint, we look at efficient semantic segmentation from a perspective of comprehensive knowledge distillation and \rev{aim} to bridge the gap between multi-source knowledge extractions and transformer-specific patch embeddings. We put forward the \emph{Transformer-based Knowledge Distillation (TransKD)} framework which learns compact student transformers by distilling both feature maps and patch embeddings of large teacher transformers, bypassing the long pre-training process and reducing the FLOPs by ${>}85.0\%$. Specifically, we propose two fundamental modules \rev{to realize feature map distillation and patch embedding distillation, respectively}: (1) \emph{Cross Selective Fusion (CSF)} enables knowledge transfer between cross-stage features via channel attention and feature map distillation within hierarchical transformers; (2) \emph{Patch Embedding Alignment (PEA)} performs dimensional transformation within the patchifying process to facilitate the patch embedding distillation. \rev{Furthermore, we introduce two optimization modules to enhance the patch embedding distillation from different perspectives: }(1) \emph{Global-Local Context Mixer (GL-Mixer)} extracts both global and local information of a representative embedding; (2) \emph{Embedding Assistant (EA)} acts as an embedding method to seamlessly bridge teacher and student models with the teacher's number of channels. 
Experiments on Cityscapes, ACDC, NYUv2, and Pascal VOC2012 datasets show that TransKD outperforms state-of-the-art distillation frameworks and rivals the time-consuming pre-training method. The source code is publicly available at \url{https://github.com/RuipingL/TransKD}.

%% file: Tex_content/introduction.tex
\begin{figure}[!t]
% \footnotesize
\small
% \scriptsize
\setlength\tabcolsep{1pt}
\centering

\begin{tabular}{c c c}
\raisebox{-0.5\height}{\includegraphics[width=0.155\textwidth]{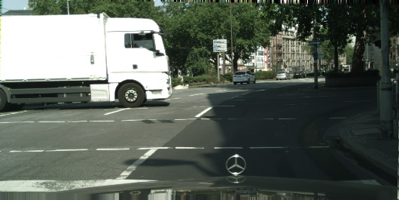}}&
\raisebox{-0.5\height}{\includegraphics[width=0.155\textwidth]{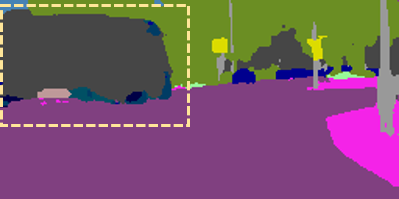}} &
\raisebox{-0.5\height}{\includegraphics[width=0.155\textwidth]{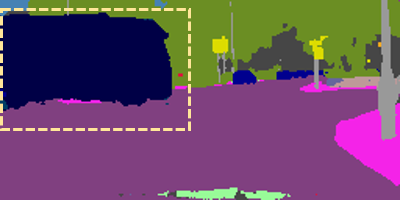}}\\
Image&{Response KD}&TransKD\\
\raisebox{-0.5\height}{\includegraphics[width=0.155\textwidth]{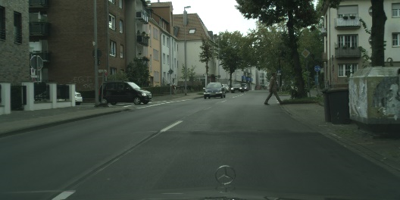}}&
\raisebox{-0.5\height}{\includegraphics[width=0.155\textwidth]{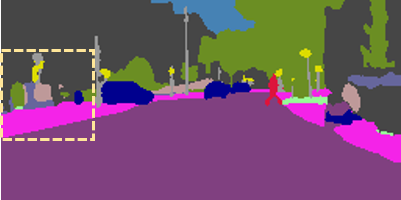}} &
\raisebox{-0.5\height}{\includegraphics[width=0.155\textwidth]{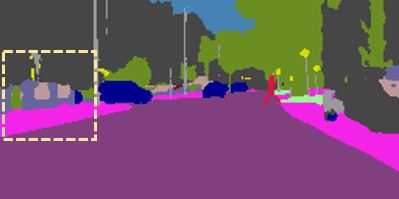}}\\
Image&{Feature KD}&TransKD
\end{tabular}
\caption{Hard examples for semantic segmentation with knowledge distillation (KD) methods. As compared to the Response-based~\cite{shu2021channel_distillation} and Feature-based~\cite{chen2021knowledge_review} KD methods, 
our framework, TransKD, enables the model to predict the \emph{truck} and \emph{fence} more precisely by exploring the knowledge from both feature maps and patch embeddings.}
\label{fig:head}
\end{figure}

\begin{figure}[!t]
\centering
\subfloat[Number of parameters\label{fig:intro_params}]{%
  \includegraphics[width=0.235\textwidth]
  {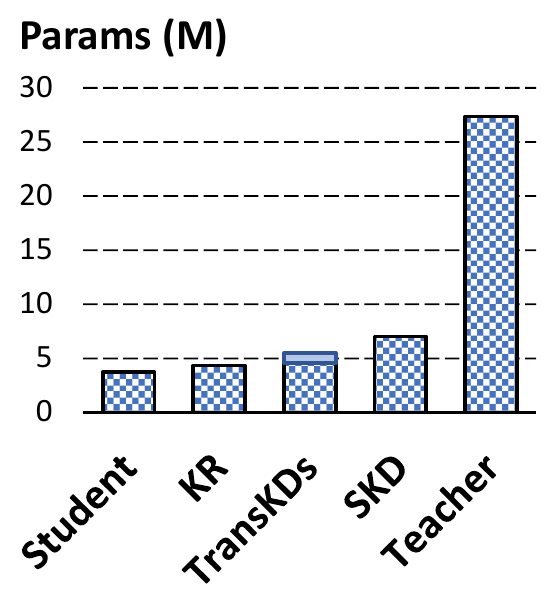}
}
\hfil
\subfloat[Performances\label{fig:intro_mIoU}]{%
  \includegraphics[width=0.235\textwidth]
  {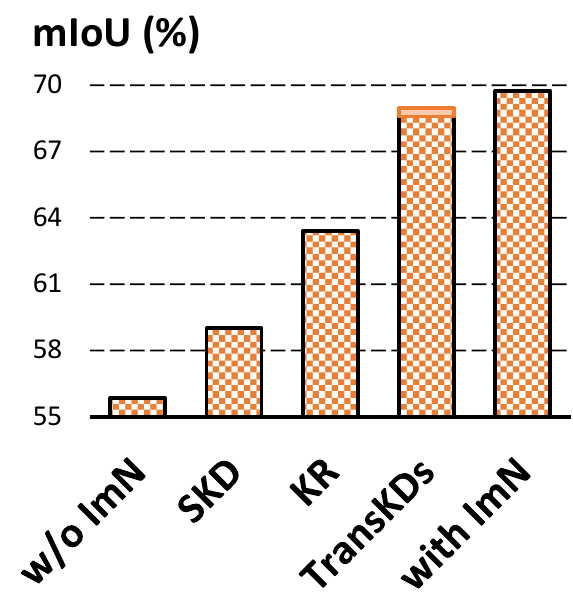}
}
\caption{Comparison with knowledge distillation frameworks. Our TransKDs compensate for the performance gap between the non-pretrained and pre-trained models effectively while adding negligible parameters. The gains over bars indicate the different amounts of parameters and performances of TransKDs. ImN: pre-training on ImageNet~\cite{deng2009imagenet}. SKD: Structured Knowledge Distillation~\cite{liu2019structured_knowledge_distillation}. KR: Knowledge Review~\cite{chen2021knowledge_review}.}
\label{fig:intro_paramsvsmIoU}
\end{figure}

\IEEEPARstart{S}{emantic} segmentation assigns category labels at a pixel-level 
(see examples in Fig.~\ref{fig:head}) 
and is a crucial tool in a wide range of applications, such as autonomous driving~\cite{romera2018erfnet,liu2022cmx,zheng2023semantic_thermal_videos,muhammad2022vision_achievements} 
and navigation assistance 
for vulnerable road users~\cite{zhang2022trans4trans,zhang2021exploring_event}.
Driven by the achievements of deep learning, semantic segmentation has empowered precise scene comprehension in the context of autonomous driving, as evidenced by remarkable accuracies~\cite{zheng2021setr, strudel2021segmenter}. 
\rev{Autonomous driving requires high-speed operation and the capacity to respond from significant distances to detect potential hazards. However, the computational resources available on vehicle mobile platforms are often insufficient to support large, effective semantic segmentation models, and compact but less effective models generally fail to distinguish distant details.} 
Moreover, there has been a notable oversight in addressing the computational limitations crucial for real-world applications, including the speed of inference and training, as well as memory footprint. 

While Convolutional Neural Networks (CNNs) have been the visual recognition front-runners for almost a decade, they encounter difficulties capturing large-scale interactions due to their focus on local neighborhood operations and limited receptive fields.
Vision transformers~\cite{zheng2021setr,strudel2021segmenter,dosovitskiy2021vit} overcome this challenge through the concept of self-attention~\cite{vaswani2017attention}, which learns to intensify or weaken parts of input at a global scale.
This excellent quality of modeling long-range dependencies 
has placed transformers at the forefront of virtually all 
segmentation benchmarks~\cite{zheng2021setr,strudel2021segmenter,dosovitskiy2021vit}. 
Unfortunately, transformers are usually very large and do not generalize well when trained on insufficient amounts of data due to the lack of inductive bias compared to CNNs~\cite{dosovitskiy2021vit}, such as translation equivariance and locality, \rev{which is relevant to the semantic segmentation task}. 
Thus, vision transformers often show less satisfactory performance on smaller task-specific datasets, and time-consuming pre-training process on a massive dataset is entailed~\cite{zheng2021setr, dosovitskiy2021vit}, \textit{e.g.}, ${\geq}100$ epochs on ImageNet~\cite{deng2009imagenet}. 
The high computational and training costs constitute a significant bottleneck for real applications, where cumbersome models cannot be deployed on mobile devices, whereas smaller models suffer from inaccurate predictions~\cite{liu2019structured_knowledge_distillation,mehta2018espnet}.

\begin{figure}
    \centering
    \includegraphics[width=0.495\textwidth]{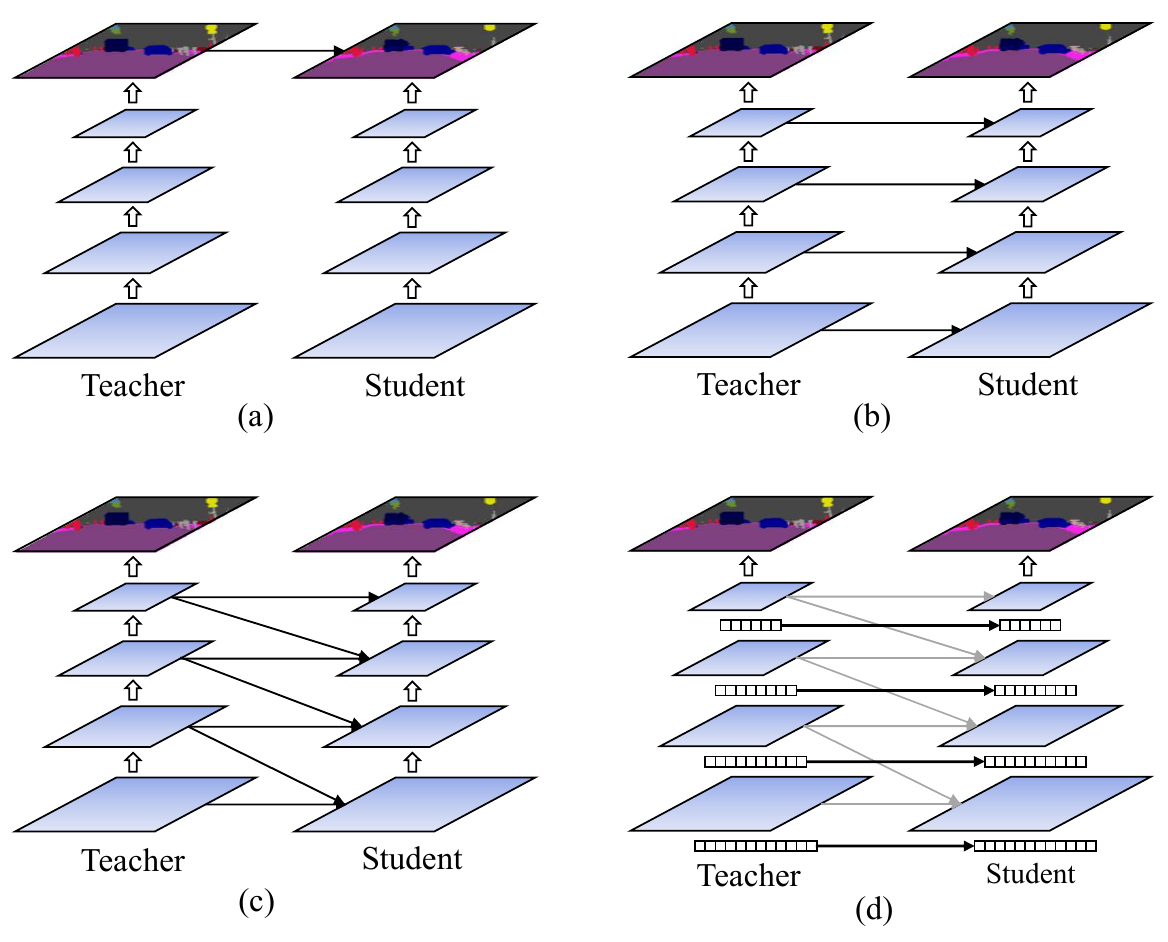}
    \caption{(a)-(c) Knowledge distillation in computer vision is split into three categories~\cite{gou2021knowledge_distillation_survey}: response-based knowledge distillation, feature-based knowledge distillation, and relation-based knowledge distillation. (d) TransKD extracts the relation-based knowledge of feature maps and transformer-specific patch embedding knowledge at each stage.}
    \label{fig:survey}
\end{figure}

A popular way to reduce the computational cost is \textit{knowledge distillation}~\cite{hinton2015distilling_knowledge}, which offers a mechanism for effective knowledge transfer between a large teacher model and a smaller student network and has been widely studied for deep CNNs~\cite{liu2019structured_knowledge_distillation,shu2021channel_distillation,chen2021knowledge_review}.
Gou~\textit{et al.}~\cite{gou2021knowledge_distillation_survey} distinguish three types of knowledge to be distilled (see Fig.~\ref{fig:survey}a-c): response-based knowledge~\cite{liu2019structured_knowledge_distillation, shu2021channel_distillation,hinton2015distilling_knowledge}, feature-based knowledge~\cite{liu2019structured_knowledge_distillation,wang2020intra_class_feature_variation_distillation,ji2022structural_statistical_texture_distillation}, and relation-based knowledge~\cite{chen2021knowledge_review,feng2021double_similarity,an2022efficient_self_distillation}. 
The response-based and feature-based knowledge distillation utilizes the output of a single specific layer of the teacher model (the last or the intermediate layer respectively), whereas the relation-based knowledge distillation tracks the relationship between multiple layers simultaneously.
These paradigms are primarily designed with the CNN-to-CNN transfer in mind~\cite{liu2019structured_knowledge_distillation,shu2021channel_distillation,chen2021knowledge_review} and do not adequately cover knowledge distillation from transformer-specific intermediate blocks. 
The patch embedding method is a transformer-specific module to partition the input images or feature maps into a sequence of patches. \rev{It is typically achieved through convolution operations with learnable parameters, rather than by reshaping.}
According to the vanilla vision transformer~\cite{dosovitskiy2021vit}, simply partitioning the input images or feature maps into large non-overlapping patches is impractical to learn the spatial relations. This limitation is addressed with optimized embedding methods,~\textit{e.g.} learnable positional embeddings~\cite{dosovitskiy2021vit}, conditional positional embeddings~\cite{chu2021twins,li2022uniformer}, shifted windows~\cite{liu2021swin}, and overlapping patch embeddings~\cite{wang2022pvt_v2,xie2021segformer}. 
Viewing that there is a large gap between the performances of the transformers with a simple patchifying process and optimized embedding methods,
\rev{we recognize the importance of the sequential information within patch embeddings, which are processed by the embedding module that transforms feature maps into these embeddings.}
Therefore, we regard patch embeddings as a transformer-specific knowledge source.

To achieve efficient semantic segmentation, we propose \emph{Transformer-based Knowledge Distillation (TransKD)}, the first framework designed to distill knowledge from large-scale semantic segmentation transformers. 
We focus on comprehensive distillation from different knowledge sources within the transformer-specific building blocks (our key idea is illustrated in Fig.~\ref{fig:survey}d).
TransKD distills the information from both (1) \textit{feature maps} and (2) the transformer-specific \textit{patch embeddings}.
As mentioned before, patch embeddings differ from
feature maps not only in the distinct shape and dimensionality but also in the inherent knowledge acquired through the patchifying process.
\rev{Therefore, we consider that transformer-specific patch embeddings should be fully exploited to maximize the potential of transformer-transformer knowledge distillation. By combining knowledge from feature maps and patch embeddings, both spatial and sequential relationships can be harvested, thereby boosting efficient semantic segmentation through distillation.}

Specifically, we design two fundamental and two optimization modules within the TransKD framework.
(1) For feature map distillation, we build a relation-based scheme atop Knowledge Review~\cite{chen2021knowledge_review} and introduce the  \emph{Cross Selective Fusion (CSF)} module to merge cross-stage feature maps via channel attention and construct the feature map distillation streams within hierarchical transformers.
(2) For patch embedding distillation, we introduce the fundamental \emph{Patch Embedding Alignment (PEA)} module to perform a dimensional transformation of patch embeddings along the channel dimension to facilitate multi-stage patch embedding distillation streams.
(3) Recognizing that both neighboring and long-range relationships are vital for semantic segmentation~\cite{xie2021segformer,zhao2017pspnet}, our \emph{Global-Local Context Mixer (GL-Mixer)} extracts both global and local information of a representative embedding for distillation.
(4) Furthermore, considering the large gap between the student's and the teacher's size that negatively impacts the knowledge distillation results~\cite{mirzadeh2020teacher_assistant,huang2022knowledge_distillation_stronger_teacher}, we present the \emph{Embedding Assistant (EA)} module, which acts an embedder with teacher's number of channels. 
EA helps to build a pseudo teacher assistant model by combining the student's transformer blocks and seamlessly bridge teacher and student models.

We demonstrate the benefits of our approach for real-world mobile applications on four benchmarks: a general street scene dataset (Cityscapes~\cite{cordts2016cityscapes}), an adverse street scene dataset (ACDC~\cite{sakaridis2021acdc}), an indoor understanding dataset (NYUv2~\cite{silberman2012nyu_dataset}), and an object-centric dataset (Pascal VOC2012~\cite{Everingham2010voc}).
Comprehensive experiments showcase that our framework outperforms state-of-the-art 
KD counterparts by a large gap.
Compared to the cumbersome teacher model, TransKD reduces the floating-point operations (FLOPs) by ${>}85.0\%$, while maintaining competitive accuracy.
By unifying patch embedding and feature map distillation, TransKD robustifies the segmentation of hard examples where previous paradigms struggle, as shown in Fig.~\ref{fig:head}.
Benchmarked against the feature-map-only method Knowledge Review~\cite{chen2021knowledge_review}, TransKD-Base enhances the distillation performance by $5.18\%$ in mean Intersection over Union (mIoU) while adding negligible $0.21M$ parameter during the training phase, as shown in Fig.~\ref{fig:intro_paramsvsmIoU}.
On Cityscapes, TransKD improves the mIoU of the non-pretrained SegFormer-B0~\cite{xie2021segformer} by $13.12\%$ and the pre-trained one by $2.09\%$.
We further validate various 
transformer models~\cite{wang2022pvt_v2, xie2021segformer,yang2022lite_vision_transformer} and conform that TransKD consistently boosts the accuracy of compact segmenters. 
Lastly, TransKD rivals the performance of the time-consuming pre-trained method.
Our best model achieves $75.74\%$ in mIoU with only $3.72M$ parameters. 

At a glance, this work delivers the following contributions:
\begin{compactitem}
    \item We are the first to propose a transformer-to-transformer knowledge distillation framework in semantic segmentation and \rev{the first} to utilize transformer-specific patch embedding as one of the knowledge sources.
    \item We provide a new perspective on distilling knowledge from transformer-specific patch embeddings. Our TransKD framework unifies feature map and patch embedding distillation.
    \item To this intent, we design two fundamental modules, \emph{Cross Selective Fusion (CSF)} and \emph{Patch Embedding Alignment (PEA)}, and two optimization modules, \emph{Global-Local Context Mixer (GL-Mixer)} and \emph{Embedding Assistant (EA)}.  
    \item In-depth experiments validate the advantages of TransKD on Cityscapes, ACDC, NYUv2, and Pascal VOC2012 datasets, demonstrating significant improvements over existing knowledge distillation methods.
\end{compactitem}

%% file: Tex_content/related_work.tex
\subsection{From Accurate to Efficient Semantic Segmentation}
Semantic segmentation has witnessed tremendous progress since Fully Convolutional Networks (FCNs)~\cite{long2015fcn} first looked at the pixel classification problem from an end-to-end perspective.
Many subsequent networks followed this scheme, often raising its accuracy through encoder-decoder architectures~\cite{badrinarayanan2017segnet} or the aggregation of multi-scale context~\cite{zhao2017pspnet,chen2018deeplabv3+}.
An important group of methods leverages non-local self-attention~\cite{vaswani2017attention,wang2018nonlocal} to capture long-range contextual dependencies and promote global reasoning~\cite{fu2019danet,yuan2021ocnet}.
The recent success of vision transformers~\cite{dosovitskiy2021vit,yu2021metaformer} leads to the development of multiple transformer-based architectures for semantic segmentation~\cite{zheng2021setr,strudel2021segmenter} which have since then become state-of-the-art on mainstream benchmarks~\cite{cordts2016cityscapes,silberman2012nyu_dataset,sakaridis2021acdc}, thanks to their ability to capture global dependencies from early layers via token mixing modules such as self-attention~\cite{vaswani2017attention} or \rev{Multi-Layer Perceptron (MLP)} blocks~\cite{tolstikhin2021mlp_mixer}.
Yet, the computational cost and lengthy training process remain a weak spot of the accuracy-driven transformer research.

On the other hand, multiple works explicitly address efficient semantic segmentation, aiming to strike a good balance between speed and accuracy.
This line of research compactifies architectures through techniques such as early downsampling~\cite{paszke2016enet}, filter factorization~\cite{romera2018erfnet,mehta2018espnet}, multi-branch architecture~\cite{zhao2018icnet,yu2018bisenet}, and ladder-style upsampling~\cite{orsic2019swiftnet,li2019dfanet}.
Lightweight classification backbones such as MobileNets~\cite{sandler2018mobilenetv2} and ShuffleNets~\cite{ma2018shufflenet_v2} 
have also been adapted to expedite image segmentation. 
Moreover, SegFormer~\cite{xie2021segformer} introduces a lightweight and efficient semantic segmentation framework that unifies the backbone output feature maps with a simple MLP decoder.
Unlike all these works, our TransKD views resource-efficient segmentation transformers from a new perspective of comprehensive knowledge distillation. We introduce a framework that simultaneously extracts knowledge from different blocks of large-scale vision transformers and reuses it for compact and fast student transformers.

\subsection{Vision Transformer for Dense Prediction}
Transformer architecture has become the de-facto standard in Natural Language Processing (NLP), mostly due to the effectiveness of stacked self-attention and its ability to capture long-range relationships within the data.
The success of self-attention-based models has also extended to the field of visual recognition, beginning with the introduction of the Vision Transformer (ViT)~\cite{dosovitskiy2021vit} -- the first transformer to deliver strong results on ImageNet~\cite{deng2009imagenet}.
Subsequent developments have further optimized the performance of the vision transformer, incorporating features such as multi-scale networks~\cite{liu2021swin,xie2021segformer,wang2021pvt}, increased depth~\cite{zhou2021deepvit,touvron2021cait}, knowledge distillation~\cite{touvron2021deit,graham2021levit} and a blend of global and local attention~\cite{chu2021twins,wang2021crossformer}.

At the same time, multiple lightweight vision transformers~\cite{graham2021levit,wu2022tinyvit} have emerged.
For example, LeViT~\cite{graham2021levit} is a hybrid model reusing ResNet stages within the transformer architecture while MobileViT~\cite{mehta2021mobilevit} leverages MoibleNetV2 blocks~\cite{sandler2018mobilenetv2} along the downsampling path and Mobile-Former~\cite{chen2021mobile_former} establishes a dual-stream architecture by bridging MobileNet and its transformer branch.
Unfortunately, recent studies~\cite{mehta2021mobilevit} suggest that MobileViT and other ViT-based models are still not efficient enough for real-time processing on mobile devices. 
In this work, we develop a knowledge distillation framework based on vision transformers, which consistently improves the performance of compact dense prediction and semantic segmentation transformers~\cite{xie2021segformer,wang2022pvt_v2,yang2022lite_vision_transformer}, leading to a much better trade-off of accuracy, computational cost, and pre-training requirements. 

\begin{figure*}[t!]
    \centering
    \includegraphics[scale=0.7]{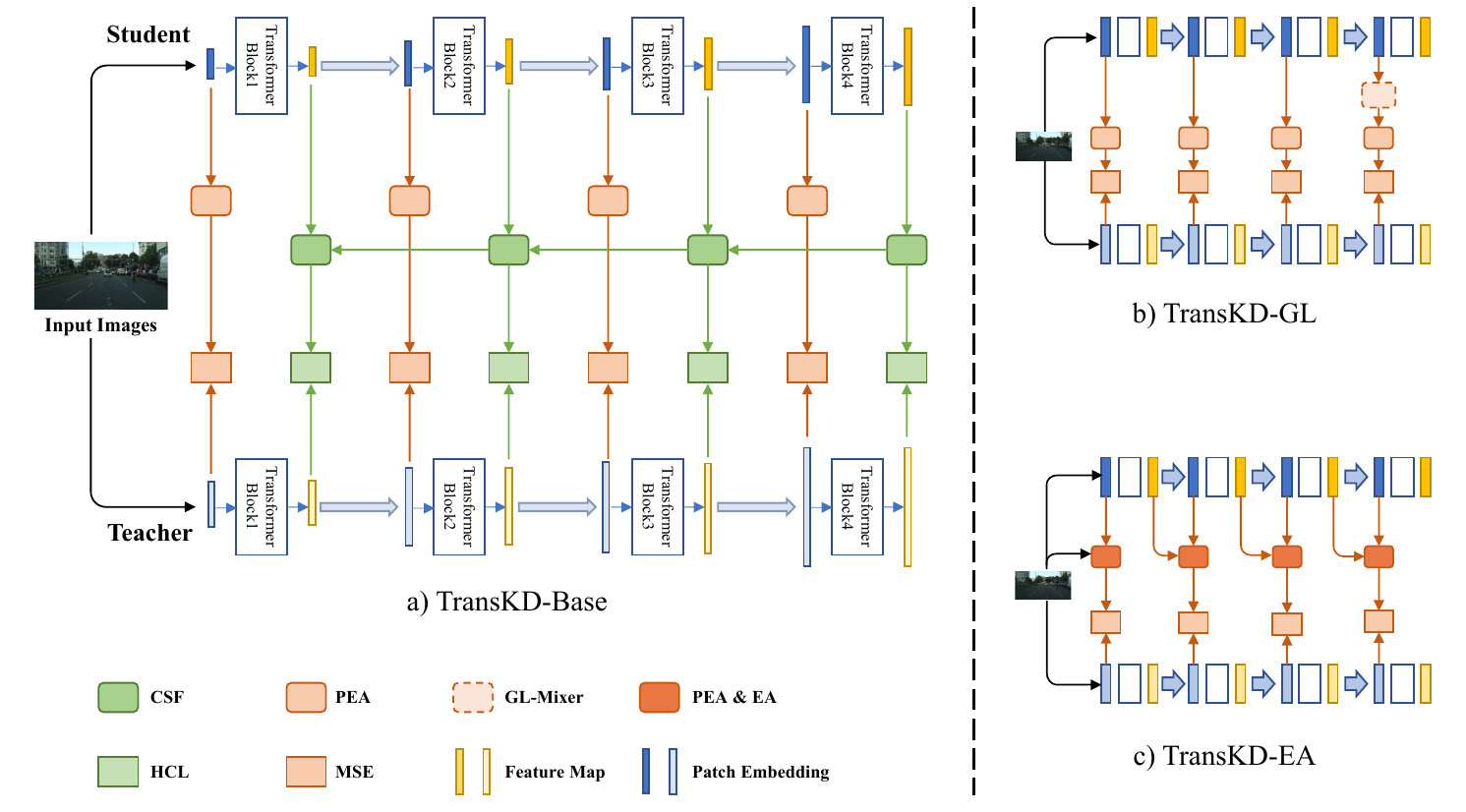}
    \caption{Our knowledge distillation framework TransKD. It is divided into two parts: knowledge distillation of patch embeddings (indicated by red arrows and rectangles) and feature maps (indicated by green arrows and rectangles).
    The loss function consists of two distillation terms (HCL and MSE) and a cross-entropy term.
    (a) TransKD-Base is the basic version of TransKD, constructed with two fundamental modules, CSF and PEA.
    (b) TransKD-GL and (c) TransKD-EA are two optimized versions of TransKD.
    }
    \label{fig:KD_structure}
\end{figure*}
\subsection{Knowledge Distillation for Semantic Segmentation}
Hinton~\textit{et al.}~\cite{hinton2015distilling_knowledge} introduced the concept of Knowledge Distillation (KD), which involves transferring knowledge from a large-scale model to a more compact one.
These two models act as teacher and student, and this paradigm is widely used in computer vision~\cite{gou2021knowledge_distillation_survey,wang2021knowledge_distillation_review}.
In this work, our goal is to study and design a KD framework specifically for semantic segmentation with transformer-based models.

KD approaches have been categorized~\cite{gou2021knowledge_distillation_survey} into three types based on the form of knowledge:
response-based KD~\cite{liu2019structured_knowledge_distillation,shu2021channel_distillation,wang2020intra_class_feature_variation_distillation},
feature-based KD~\cite{shu2021channel_distillation,wang2020intra_class_feature_variation_distillation,ji2022structural_statistical_texture_distillation,liu2021exploring_inter_channel_correlation,yang2022cross_relational_distillation,zhang2022distilling_inter_class_distance},
and relation-based KD~\cite{chen2021knowledge_review,feng2021double_similarity,an2022efficient_self_distillation}.
Liu~\textit{et al.}~\cite{liu2019structured_knowledge_distillation} first introduced the idea of structured knowledge distillation, leveraging adversarial learning to align the segmentation map produced by the compact student network with the one of the cumbersome teacher network.
Knowledge Review (KR)~\cite{chen2021knowledge_review} for the first time proposed to distill knowledge using cross-stage connection paths.
Double Similarity Distillation (DSD)~\cite{feng2021double_similarity} transfers both detailed spatial dependencies and global category correlations.
Unlike previous works primarily devoted to the distillation of spatial-wise knowledge, some recent works focus on the distillation of channel-wise distribution.
Channel Distillation (CD)~\cite{shu2021channel_distillation} emphasizes the soft distributions of channels and pays attention to the most salient parts of the channel-wise maps.

Additionally, knowledge adaptation is addressed in~\cite{he2019knowledge_adaptation} by optimizing affinity distillation. Adaptive Perspective Distillation (APD)~\cite{tian2022adaptive_perspective_distillation} mines detailed contextual cues from each training sample, whereas~\cite{lin2022knowledge_distillation_target_aware_transformer} employs a hierarchical distillation approach and enables one-to-all spatial matching.
All these works are dedicated to CNN-to-CNN or CNN-to-transformer knowledge distillation.
In contrast, our proposed TransKD model explores this concept in a novel transformer-to-transformer fashion, by explicitly considering multi-source knowledge within the transformer architecture and taking transformer-specific patch embeddings into account.

%% file: tables/configuration.tex
\begin{tabular}{c|c|l|c}
\toprule
         \textbf{Framework} & \textbf{FM Distillation} & \textbf{PE Distillation}& \textbf{Plug\&Play} \\
         \midrule
         TransKD-Base&CSF&PEA&{\Checkmark}\\
         TransKD-GL&CSF&PEA + GL-Mixer&{\Checkmark}\\
         TransKD-EA&CSF&PEA + EA&{\XSolid}\\
\bottomrule
\end{tabular}

%% file: Tex_content/experiment_datasets_and_setups.tex
\subsection{Datasets and Metrics}
\noindent\textbf{Cityscapes}~\cite{cordts2016cityscapes} is a large-scale dataset, which contains sequences of street scenes from $50$ different cities.
Cityscapes consists of $2975$ training- and $500$ validation images with dense annotations as well as $1525$ test images.
It is well known for assessing the performance of vision algorithms for urban semantic scene understanding.
There are $19$ semantic classes.

\noindent\textbf{ACDC}~\cite{sakaridis2021acdc} is a dataset used to address semantic segmentation under adverse conditions, \textit{e.g.}, fog, nighttime, rain, and snow scenes.
Each adverse condition includes $400$ training, $100$ validation, and $500$ test images, except for the nighttime condition which includes $106$ validation images.
Tougher and more comprehensive situations are characteristic of the ACDC dataset, as its goal is to validate the quality of semantic segmentation under adverse conditions with the identical label set of $19$ classes as Cityscapes.

\noindent\textbf{NYU Depth V2}~\cite{silberman2012nyu_dataset} is an indoor dataset captured by both the RGB and depth cameras.
It contains scenes of offices, stores, and rooms of houses.
There are many occluded objects with uneven illumination.
The dataset includes $1449$ RGB-depth images, divided into $795$ training images and $654$ testing images.
The images are annotated with $40$ semantic categories. We use RGB images in our experiments investigating knowledge distillation.

\noindent\textbf{Pascal VOC2012}~\cite{Everingham2010voc} is a dataset for object recognition competitions. The images are annotated with $20$ foreground object categories and one background class. We adopt the augmentation strategy in~\cite{Hariharan2011vocaug}. The augmented large-scale dataset contains $10582$/$1449$/$1456$ images for training/validation/testing.

\noindent\textbf{Evaluation metric.} The mean Intersection over Union (mIoU) over all classes is a common evaluation metric for semantic segmentation tasks. All experiments are evaluated based on mIoU, which is calculated via Eq.~(\ref{eq:miou}):
\begin{equation}\label{eq:miou}
mIoU=\frac{1}{k+1}\sum_{i{=}0}^{k}\frac{p_{ii}}{\sum_{j=0}^{k}p_{ij}+\sum_{j=0}^{k}\left(p_{ji}-p_{ii}\right)},
\end{equation}
where $k$ is the number of classes, $p_{ij}$ is the number of pixels belonging to the class $i$ and classified as class $j$.
\rev{To evaluate the efficiency of TransKDs,} we use MMCV\footnote{MMCV: https://github.com/open-mmlab/mmcv} to compute \rev{Floating point operations per second (FLOPs)} and the number of parameters of our models, \rev{
Additionally, we use the source code to test the Frames Per Second (FPS) of the models on an NVIDIA GeForce GTX 1080 Ti GPU.
}.

\subsection{Implementation Details} 
\noindent\textbf{Teacher and student transformer models.}
For the main part of our semantic segmentation experiments, 
we investigate knowledge distillation for compressing a large pre-trained SegFormer B2 to a smaller SegFormer B0 model without pre-training (unless specified), which has a similar number of parameters as ResNets used in previous distillation works~\cite{liu2019structured_knowledge_distillation,shu2021channel_distillation,chen2021knowledge_review}.
SegFormer B0~\cite{xie2021segformer} achieves a speed of $35.06$ Frames Per Second (FPS) when running at a resolution of $512{\times}1024$ on e NVIDIA GTX 1080Ti and is therefore well-suited for real-time applications.
We have also examined our framework with PVTv2~\cite{wang2022pvt_v2} and LVT~\cite{yang2022lite_vision_transformer} to assess the generality of our approach.

\noindent\textbf{Training setups and hyperparameters.}
When evaluating the performance of the knowledge distillation frameworks, we resize the images to $512{\times}1024$ for  Cityscapes~\cite{cordts2016cityscapes}, 
$512{\times}910$ for ACDC~\cite{sakaridis2021acdc}, and $512{\times}512$ for Pascal VOC2012~\cite{Everingham2010voc},
while maintaining the original image scale of $480{\times}640$ for the NYUv2 dataset~\cite{silberman2012nyu_dataset}.
We utilize one NVIDIA GTX 1080Ti GPU when experimenting with the aforementioned input sizes and four NVIDIA A100 GPUs for larger input sizes.
The number of epochs is set to be $1000$ on Cityscapes, ACDC, and Pascal VOC2012 datasets, and $1500$ on the NYUv2 dataset. 
Following the structure of SegFormer~\cite{xie2021segformer}, the height and width of the output and the target segmentation maps are all rescaled with a ratio of $1/4$ compared to the input image.
The performance of the baseline SegFormer model has a slight decay, which is reasonable due to the smaller size of the input image. 
The models in our work are trained using the AdamW optimizer~\cite{loshchilov2017adamw} 
with the learning rate of $6{\times}10^{-5}$, \rev{and the default epsilon of $1\times 10^{-8}$ and betas $(0.9, 0.999)$}. \rev{The learning rate is adjusted using a polynomial learning rate decay scheduler~\cite{Mishra2019polylr} with a default factor of $1.0$. The end learning rate is set to $0.0$, and the maximum number of decay steps is set at $1500$.}
We adopt a batch size of $8$ \rev{per GPU} on VOC2012 and $2$ on other datasets.
Additionally, the number of channels of the feature maps in the CSF fusion module $C$ is set to $64$.

%% file: tables/comparison_with_KD_methods.tex
\begin{tabular}{c|c|c|l}
\toprule
\textbf{Network} & \textbf{\#Params (M)} & \textbf{GFLOPs} & \textbf{mIoU (\%)}\\
\midrule
Teacher (B2) &27.36 &113.84&76.49\\
Student (B0)&3.72&13.67&55.86\\
\midrule
+Pre-train&3.72&13.67&69.75\\
\midrule
+KD~\cite{hinton2015distilling_knowledge}&3.72&13.67&56.89\\
+CD~\cite{shu2021channel_distillation}&3.72&13.67&61.90\\
+SKD (PA)~\cite{liu2019structured_knowledge_distillation}&3.85&13.74&58.05\\
+SKD (HO)~\cite{liu2019structured_knowledge_distillation}&7.04&13.83&59.00\\
+Knowledge Review*~\cite{chen2021knowledge_review}&4.35&16.17&62.41\\
+Knowledge Review~\cite{chen2021knowledge_review}&4.35&16.17&63.40\\
+TransKD-Base (ours)&4.56&16.47&68.58~\gbf{+5.18}\\
+TransKD-GL (ours)&5.22&16.80&68.87~\gbf{+5.47}\\
\textbf{+TransKD-EA (ours)}&5.53&17.84&\textbf{68.98}~\gbf{+5.58}\\
\bottomrule
\end{tabular}

%% file: tables/cityscapes_class.tex
	\begin{tabular}{c|c|c|c|c|c|c|c|c|c|c}
   \toprule
   \textbf{Method} &\textbf{mIoU}     &\textbf{road}  &\textbf{sidewalk}     &\textbf{building}      &\textbf{wall}      &\textbf{fence}      &\textbf{pole}   &\textbf{traffic light}      &\textbf{traffic sign}      &\textbf{vegetation} \\
   \midrule
   Knowledge Review~\cite{chen2021knowledge_review}  &63.40       &97.09      &78.07         &88.26    &37.55      &44.12    &46.55      &53.15              &60.96             &89.39 \\
         TransKD-Base    &68.58  &97.47 &80.29    &89.71       &47.13      &{\bf48.25} &53.78     &56.24        &65.82 &90.74 \\
      TransKD-GL    &68.87  &{\textbf{97.59}} &{\textbf{80.50}}    &{\textbf{89.83}}       &47.87      &46.80 &{\textbf{54.34}}     &{\textbf{57.97}}        &{\textbf{66.65}} &{\textbf{90.88}} \\
      \textbf{TransKD-EA}    &{\textbf{68.98}}  &97.46 &79.91    &89.75       &{\textbf{54.37}}      &47.87 &53.21     &56.06        &65.25 &90.70 \\
  \midrule
   \textbf{Class} &\textbf{terrain}      &\textbf{sky}        &\textbf{person}      &\textbf{rider}      &\textbf{car}        &\textbf{truck}      &\textbf{bus}        &\textbf{train}      &\textbf{motorcycle}      &\textbf{bicycle} \\
\midrule
   Knowledge Review~\cite{chen2021knowledge_review}  &57.74        &92.34      &65.26       &40.58      &89.83      &50.92      &64.30      &44.29      &40.79           &63.40 \\
         TransKD-Base  &60.63   &{\bf93.05} &71.92  &45.76 &91.84 &66.02 &{\bf74.01} &63.31 &39.60           &67.54 \\
      TransKD-GL   &{\textbf{62.07}}   &92.73 &{\textbf{72.21}}  &{\textbf{46.15}} &{\textbf{91.88}} &{\textbf{70.37}} &67.39 &57.18 &\textbf{48.28}           &{\textbf{67.81}} \\
      \textbf{TransKD-EA}   &60.94   &92.59 &71.14  &45.89 &91.61 &65.56 &71.17 &{\textbf{66.96}} &43.46         &66.68 \\
\bottomrule
	\end{tabular}

%% file: tables/various_architectures.tex
\begin{tabular}{l|c|c|c|l}
    \toprule
    \textbf{Network}&\textbf{\#Params (M)}&\textbf{GFLOPs}&\rev{\textbf{FPS}}&\textbf{mIoU (\%)}\\
    \midrule
         T: SegFormer-B2 &27.36&113.84&\rev{10.17}& 76.49\\
         S: SegFormer-B0 &3.72&13.67&\rev{35.06}& 55.86\\
         +KR &4.35&16.17&\rev{35.06}& 63.40\\
         +TransKD-Base&4.56&16.47&\rev{35.06}& \textbf{68.58}~\gbf{+5.18}\\
         \midrule
         T: PVTv2-B2 &29.1&83.14&\rev{9.91}& 77.26 \\
         S: PVTv2-B0 &7.53&47.45&\rev{24.25}& 58.35\\
         +KR &8.15&49.94&\rev{24.25}& 66.51\\
         +TransKD-Base&8.37&50.24&\rev{24.25}& \textbf{68.69}~\gbf{+2.18}\\
         \midrule
         T: SegFormer-B2 &27.36&113.84&\rev{10.17}& 76.49\\
         S: LVT &3.84&18.06&\rev{20.42}& 57.49\\
         +KR &4.47&20.63&\rev{20.42}& 65.83 \\
         +TransKD-Base&4.69&20.99&\rev{20.42}& \textbf{68.54}~\gbf{+2.71}\\
         \midrule
    \end{tabular}

%% file: tables/acdc.tex
\begin{tabular}{c|c c c c|l}
\toprule
\textbf{Network} & \textbf{Fog} & \textbf{Night} & \textbf{Rain} & \textbf{Snow} & \textbf{All-ACDC}\\
\midrule
Teacher (B2) &73.26 &51.44 &69.26 &72.00&69.34\\
Student (B0) &52.10&33.33&45.79&48.38&46.26\\
\midrule
CD~\cite{shu2021channel_distillation} &57.07&40.82&51.37&52.15&52.36\\
Knowledge Review~\cite{chen2021knowledge_review}&59.83&42.38&52.30&55.69&54.90\\
TransKD-Base (ours)&63.74&44.63&57.16&60.30&58.56~\gbf{+3.66}\\
TransKD-GL (ours)&63.18&43.55&55.85&\textbf{60.61}&58.13~\gbf{+3.23}\\
\textbf{TransKD-EA (ours)}&{\textbf{66.13}}&{\textbf{44.99}}&\textbf{57.24}&59.88&{\textbf{59.09}}~\gbf{+4.19}\\
\midrule
\end{tabular}

%% file: tables/nyu.tex
\begin{tabular}{c|c|c|l}
    \toprule
    \textbf{Network}& \textbf{\#Params (M)} & \textbf{GFLOPs} & \textbf{mIoU (\%)}\\
    \midrule
     Teacher  (B2)& 27.38 & 66.73 & 44.11\\
     Student (B0) &3.72 & 13.85 & 18.19\\
     \midrule
     +CD~\cite{shu2021channel_distillation}&3.72&13.85&20.83\\
     +Knowledge Review~\cite{chen2021knowledge_review}&4.35&16.35&22.80\\
    TransKD-Base (ours)&4.57&16.64&23.61~\gbf{+0.81}\\
    \textbf{TransKD-GL (ours)}&5.22&16.98&\textbf{24.20}~\gbf{+1.40}\\
    TransKD-EA (ours)& 5.54&18.02&23.89~\gbf{+1.09}\\
\bottomrule
\end{tabular}

%% file: tables/voc2012.tex
\begin{tabular}{c|c|c|l}
\toprule
\textbf{Network} & \textbf{\#Params (M)} & \textbf{GFLOPs} & \textbf{mIoU (\%)}\\
\midrule
Teacher (B2) &27.36 &56.95&77.71\\
Student (B1)&13.68&13.32&47.02\\
\midrule
+CD~\cite{shu2021channel_distillation}&13.68&13.32&53.94\\
+KR~\cite{chen2021knowledge_review}&14.34&14.63&61.64\\
+TransKD-Base&14.75&14.93&65.81~\gbf{+4.17}\\
+TransKD-GL&17.38&15.60&66.13~\gbf{+4.49}\\
\textbf{+TransKD-EA}&16.68&16.15&\textbf{66.32}~\gbf{+4.68}\\
\midrule
+ImN&13.68&13.32&70.82\\
+KR~\cite{chen2021knowledge_review}+ImN&14.34&14.63&73.26\\
+TransKD-Base+ImN&14.75&14.93&\textbf{73.67}\\
\bottomrule
\end{tabular}

%% file: tables/fusion_module.tex
\begin{tabular}{c|c|c|l}
\toprule
\textbf{Feature Map Fusion} & \textbf{Loss Function} &\textbf{Dataset}& \textbf{mIoU (\%)}\\
\midrule
ABF~\cite{chen2021knowledge_review}&HCL&Cityscapes&63.40\\
\textbf{CSF (ours)}&HCL&Cityscapes&\textbf{65.94}~\gbf{+2.54}\\
\midrule
ABF~\cite{chen2021knowledge_review}&HCL&ACDC&54.90\\    
\textbf{CSF (ours)}&HCL&ACDC&\textbf{55.35}~\gbf{+0.45}\\
\bottomrule
\end{tabular}

%% file: tables/loss_function.tex
\begin{tabular}{c|c|c|c}
\toprule
\textbf{Loss Function}&\textbf{Feature Map Fusion} & \textbf{PEA} & \textbf{mIoU (\%)}\\
\midrule
Channel-wise KL &CSF&All &66.80\\
Spatial-wise KL &CSF&All&66.28\\
\textbf{HCL}&CSF&All&{\bf68.58}\\
\bottomrule
\end{tabular}

%% file: tables/PEA_by_stage.tex
\begin{tabular}{c|c c c|c}
\toprule
\multirow{2}{*}{\textbf{Feature Map Distillation}}&\multirow{2}{*}{\textbf{PEA Stage}} & \multicolumn{2}{c|}{\textbf{Channels}} & \multirow{2}{*}{\textbf{mIoU (\%)}}\\
&&\textbf{Student} & \textbf{Teacher}&\\
\midrule
\multirow{6}{*}{Knowledge Review~\cite{chen2021knowledge_review}}& None & -& - &63.40\\
&1st &32& 64& 64.78\\
&2nd &64&128&65.62\\
&3rd &160&320&66.37\\
&4th &256&512&66.52\\
&All &-&-&{\bf67.89}\\
\midrule
\multirow{6}{*}{CSF+HCL (TransKD)}& None & -& - &65.94\\
&1st&32&64& 66.75\\
&2nd&64&128& 67.05\\
&3rd&160&320& 67.08\\
&4th &256&512&67.73\\
&All &-&-&{\bf68.58}\\
\bottomrule
\end{tabular}

%% file: tables/comparison_op_versions.tex
\begin{tabular}{c|c|c|c|c}
\toprule
Network & Cityscapes & ACDC & NYUv2 & VOC2012\\
\midrule
     Student&55.86&46.26&18.19&47.02\\
     KR&63.40&54.90&22.80&61.64\\
     TransKD-Base&68.58&\underline{58.56}&23.61&65.81\\
     TransKD-GL&\underline{68.87}&58.13&\bf24.20&\underline{66.13}\\
     TransKD-EA&\bf68.98&\bf59.09&\underline{23.89}&\bf66.32\\
     \bottomrule
\end{tabular}

%% file: tables/fullsize.tex
\begin{tabular}{c|l|l}
    \toprule
    \textbf{Input Size}&\textbf{Network}&\textbf{mIoU (\%)}\\
    \midrule
         \multirow{7}{*}{$512{\times}1024$}&Teacher (SegFormer-B2)&  76.49\\
         &Student (SegFormer-B0)&55.86\\
        \cmidrule{2-3}
         &+KR&63.40\\
         &+TransKD-Base (ours)& \textbf{68.58}~\gbf{+5.18}\\
         \cmidrule{2-3}
         &+ImN&69.75\\
         &+KR+ImN&71.63\\
         &+TransKD-Base+ImN& \textbf{71.84}\\%
         \midrule
        \multirow{7}{*}{$768{\times}1536$}&Teacher (SegFormer-B2)&  78.50\\
         &Student (SegFormer-B0)&59.61\\
        \cmidrule{2-3}
         &+KR&67.97\\
         &+TransKD-Base (ours)& \textbf{70.93}~\gbf{+2.96}\\
         \cmidrule{2-3}
         &+ImN&73.07\\
         &+KR+ImN& 73.84 \\
         &+TransKD-Base+ImN& \textbf{74.06}\\
         \midrule
        \multirow{7}{*}{$1024{\times}2048$}&Teacher (SegFormer-B2)&  79.90\\
         &Student (SegFormer-B0)& 61.86\\
        \cmidrule{2-3}
         &+KR&69.06\\
         &+TransKD-Base (ours)& \textbf{71.59}~\gbf{+2.53}\\
         \cmidrule{2-3}
          &+ImN&73.77\\
         &+KR+ImN&75.55\\
         &+TransKD-Base+ImN&\textbf{75.74}\\
         \bottomrule
    \end{tabular}

%% file: tables/comparison_other_result.tex
\begin{tabular}{c|c|c|c}
    \toprule
    \textbf{Network} & \textbf{\#Params (M)}&\textbf{GFLOPs}&\textbf{mIoU (\%)}\\
    \midrule
    \multicolumn{4}{c}{\textbf{Large Semantic Segmentation Models}}\\
    \midrule
    PSPNet~\cite{zhao2017pspnet} & 68.10 & - & 78.50 \\
    DeepLabV3+~\cite{chen2018deeplabv3+} &62.70 & - & 80.90 \\
    EncNet~\cite{zhang2018context_encoding} & 55.10 & - & 76.90 \\
    DANet~\cite{fu2019danet}&45.16 & 397.40 & 78.32 \\
    CCNet~\cite{huang2019ccnet} & 68.90 & - & 80.20 \\
    HANet~\cite{choi2020cars_hanet} & 65.40 & 2138.02 & 80.29\\
    OCRNet~\cite{yuan2020object_contextual} & 70.50 & - & 81.10 \\
    SETR-MLA~\cite{zheng2021setr} & 92.59 & - &78.98\\
    SETR-PUP~\cite{zheng2021setr} & 97.64 & - &79.45\\
    MaskFormer~\cite{cheng2021maskformer} & 60.00 & - & 78.50\\
    SegFormer-B4~\cite{xie2021segformer} & 64.10 & 1240.60 & 82.30 \\
    SegFormer-B5~\cite{xie2021segformer} & 84.70 & 1460.40 & 82.40 \\
    \midrule
    \multicolumn{4}{c}{\textbf{Efficient Semantic Segmentation Models}}\\
    \midrule
    ERFNet~\cite{romera2018erfnet}* & 20.00 & 27.70 & 71.50 \\
    ICNet (Res18)*~\cite{zhao2018icnet} & 26.50 & 28.30 & 67.70\\
    ESPNet~\cite{mehta2018espnet}&0.36&4.42&61.40\\
    BiSeNet (Res18)*~\cite{yu2018bisenet} & 49.00 & - &74.80\\
    ContextNet~\cite{poudel2018contextnet} & 0.85 & - &65.90\\
    SwiftNet (Res18)*~\cite{orsic2019swiftnet}&11.80 & 26.00 & 70.20\\
    CGNet~\cite{wu2020cgnet} & 0.50 & 6.00 & 63.50\\
    Fast-SCNN*~\cite{poudel2019fastscnn} & 1.11 & - & 69.15\\
    MobileNetV2*~\cite{sandler2018mobilenetv2}&14.80&142.74&73.93\\
    ShuffleNetV2*~\cite{ma2018shufflenet_v2}&12.60&117.09&70.85\\
    EfficientNet-B0~\cite{tan2019efficientnet}&4.19&7.97&58.37\\
    SegFormer-B0 (our student) & 3.72 &  54.70 & 61.86 \\
    SegFormer-B0* (our student) & 3.72 &  54.70 & 73.77\\
    \midrule
    \multicolumn{4}{c}{\textbf{Distillation Methods on Students without using Pretrained Weights}}\\
    \midrule
    SKD (PSPNet18)~\cite{liu2019structured_knowledge_distillation} & 15.24 & 128.20 & 63.20 \\
    IFVD (PSPNet18)~\cite{wang2020intra_class_feature_variation_distillation} & \textbf{3.27} & \textbf{31.53} & 63.35 \\
    CD (PSPNet18)~\cite{shu2021channel_distillation} & 13.07 & 125.80 & 71.03 \\
    IDD (PSPNet18)~\cite{zhang2022distilling_inter_class_distance} & \textbf{3.27} & \textbf{31.53} & 69.76\\
    TransKD-Base (SegFormer B0) & 3.72 & 54.70 & \textbf{71.59}\\
    \midrule
    \multicolumn{4}{c}{\textbf{Distillation Methods on Students with Pretrained Weights}}\\
    \midrule
    MD*~\cite{xie2018improving_md} & 14.35 & 64.48 & 71.90\\
    SKD (PSPNet18)*~\cite{liu2019structured_knowledge_distillation} & 15.24 & 128.20 & 72.70\\
    KA (PSPNet18)*~\cite{he2019knowledge_adaptation} & 16.31&148.20& 74.59\\
    IFVD (PSPNet18)*~\cite{wang2020intra_class_feature_variation_distillation}&13.07&125.80&74.54\\
    DSD (PSPNet18)*~\cite{feng2021double_similarity} &15.20 & 56.90 & 73.21\\
    SASD (PSPNet)*~\cite{an2022efficient_self_distillation} & 11.38 & 117.26 & 73.10 \\
    CD (PSPNet18)*~\cite{shu2021channel_distillation} & 13.07 & 125.80 & 74.58\\
    APD (PSPNet18)*~\cite{tian2022adaptive_perspective_distillation}&16.31&148.20&75.68\\
    SSTKD (PSPNet18)*~\cite{ji2022structural_statistical_texture_distillation}&13.07&125.80&75.15\\
    CIRKD (PSPNet18)*~\cite{yang2022cross_relational_distillation}&12.90&507.40&74.73\\
    TransKD-Base (SegFormer B0)* & \textbf{3.72}& \textbf{54.70} & \textbf{75.74} \\
    \bottomrule
    \end{tabular}

%% file: Tex_content/conclusion.tex
In this work, our focus lies on enhancing semantic segmentation efficiency within the context of autonomous driving through transformer-based knowledge distillation. We introduce the TransKD architecture, a pioneering transformer-to-transformer knowledge distillation framework that capitalizes on transformer-specific patch embeddings as a critical source of knowledge. TransKD learns efficient compact segmenters via feature map distillation and patch embedding distillation, 
with two fundamental modules, \textit{i.e.,} CSF and PEA, and two optimization techniques, \textit{i.e.,} GL-Mixer and EA.
To illustrate the performance of the proposed TransKD method, in-depth evaluations are conducted on Cityscapes, ACDC, NYUv2, and Pascal VOC2012 datasets with TransKD yielding state-of-the-art performance compared to the existing knowledge distillation approaches.
The experiments yield compelling evidence that the proposed transformer-specific patch embedding-level modules serve as potent sources for knowledge distillation in semantic segmentation, showcasing high potential as the cornerstone for resource-efficient yet accurate models in the domain of autonomous scene understanding.

%% file: Tex_content/supp.tex
\appendices
\section{Knowledge Distillation Loss Functions}
\noindent\textbf{Feature map loss function.} Both patch embedding and feature map losses obtained from all four stages are weighted and added to the original cross-entropy loss. For feature map distillation, we compute the HCL loss~\cite{chen2021knowledge_review} between the outputs of CSF and the teacher transformer block at each stage.
The feature map obtained from CSF at each stage with shape $\left[n,c,h,w\right]$ is divided into $4$-level context information via adaptive average pooling. The height of the resulting multi-level feature maps at one stage is $\left[h,4,2,1\right]$, which denotes that the shapes of the four abstract feature maps are $\left[n,c,h,w\right]$, $\left[n,c,4,4\right]$, $\left[n,c,2,2\right]$, and $\left[n,c,1,1\right]$.
At each stage, $L_2$ distances are utilized to distill knowledge between the context levels. 
The final HCL loss between the feature maps at the $m$-th stage is computed as:
\begin{equation}\label{eq:hcl_supp}
L_{fm}^m = {\frac{1}{1+{\sum_l}{\frac{1}{2^l}}}}{\sum_{l=0}^3}{\frac{1}{2^l}} MSE(\mathbf{F}_{m,l}^{out},\mathbf{F}_{m,l}^T),
\end{equation}
where $l$ refers to the number of level at one stage.
The purpose of placing the HCL loss after every stage is to encourage knowledge distillation at different levels of abstraction. 

\section{More Performance Analyses}
\begin{figure}[h]
    \centering
    \includegraphics[width=0.48\textwidth]{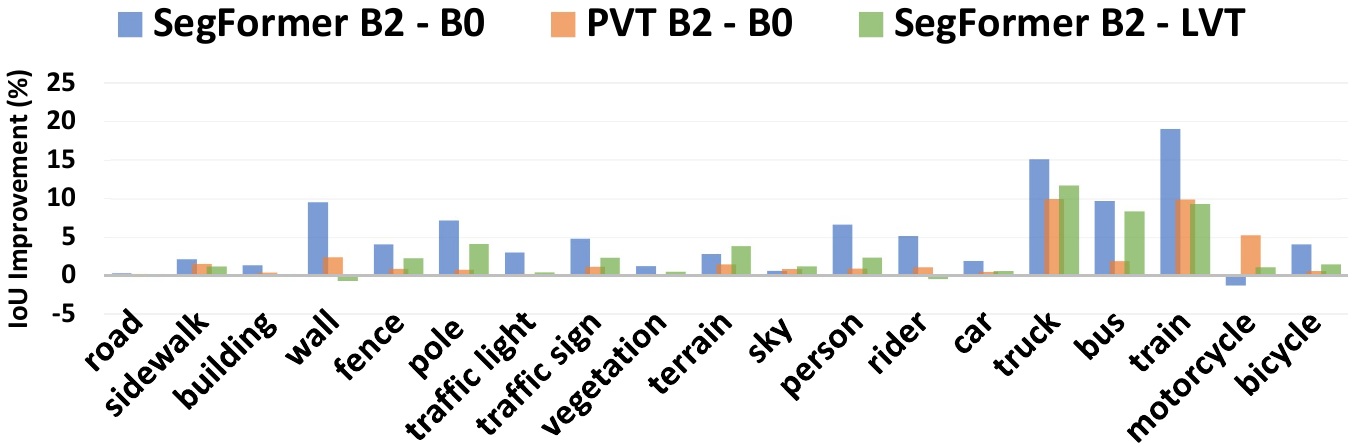}
    \caption{Per-class IoU improvements of TransKD over Knowledge Review~\cite{chen2021knowledge_review} with different teacher-student pairs.}
    \label{fig:mIoU_over_KR}
\end{figure}

\noindent\textbf{Analyses of per-class IoU improvement on various architectures.} To further investigate the effectiveness of patch embedding distillation on various architectures, Fig.~\ref{fig:mIoU_over_KR} illustrates per-class IoU improvements of our TransKD over Knowledge Review with all three teacher-student pairs, \textit{i.e.} SegFormer B2-B0~\cite{xie2021segformer}, PVT B2-B0~\cite{wang2022pvt_v2}, SegFormer B2-LVT~\cite{yang2022lite_vision_transformer}. Compared to Knowledge Review, TransKD helps the compact models to recognize less-frequent vehicle classes more accurately, such as \emph{truck}, \emph{bus}, and \emph{train}.
\begin{figure*}[!t]
\footnotesize
% \small
% \scriptsize
\setlength\tabcolsep{1pt}
\centering

\begin{tabular}{c c c c c c c}
\raisebox{-0.5\height}{\includegraphics[width=0.16\textwidth]{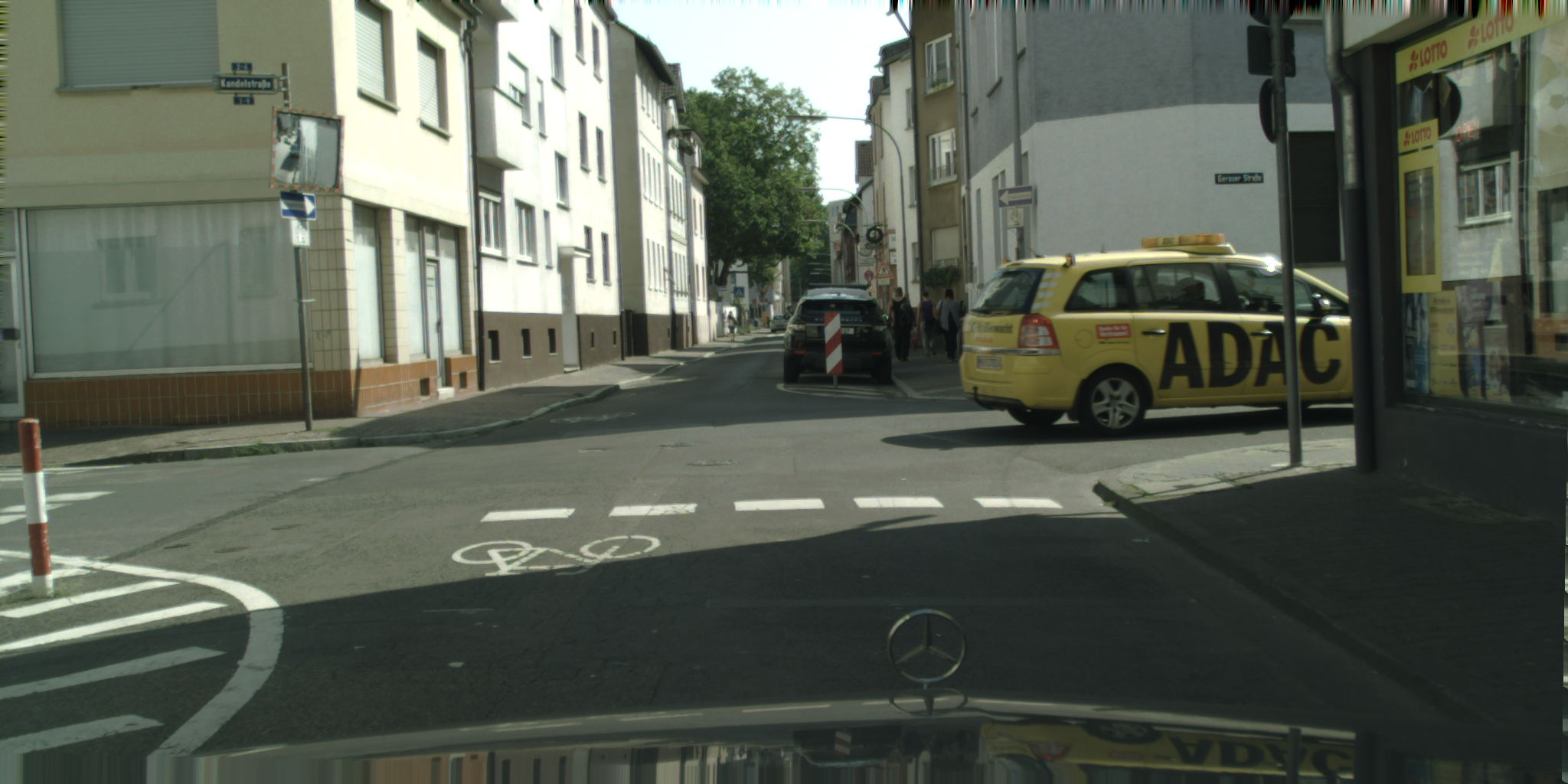}} &
\raisebox{-0.5\height}{\includegraphics[width=0.16\textwidth]{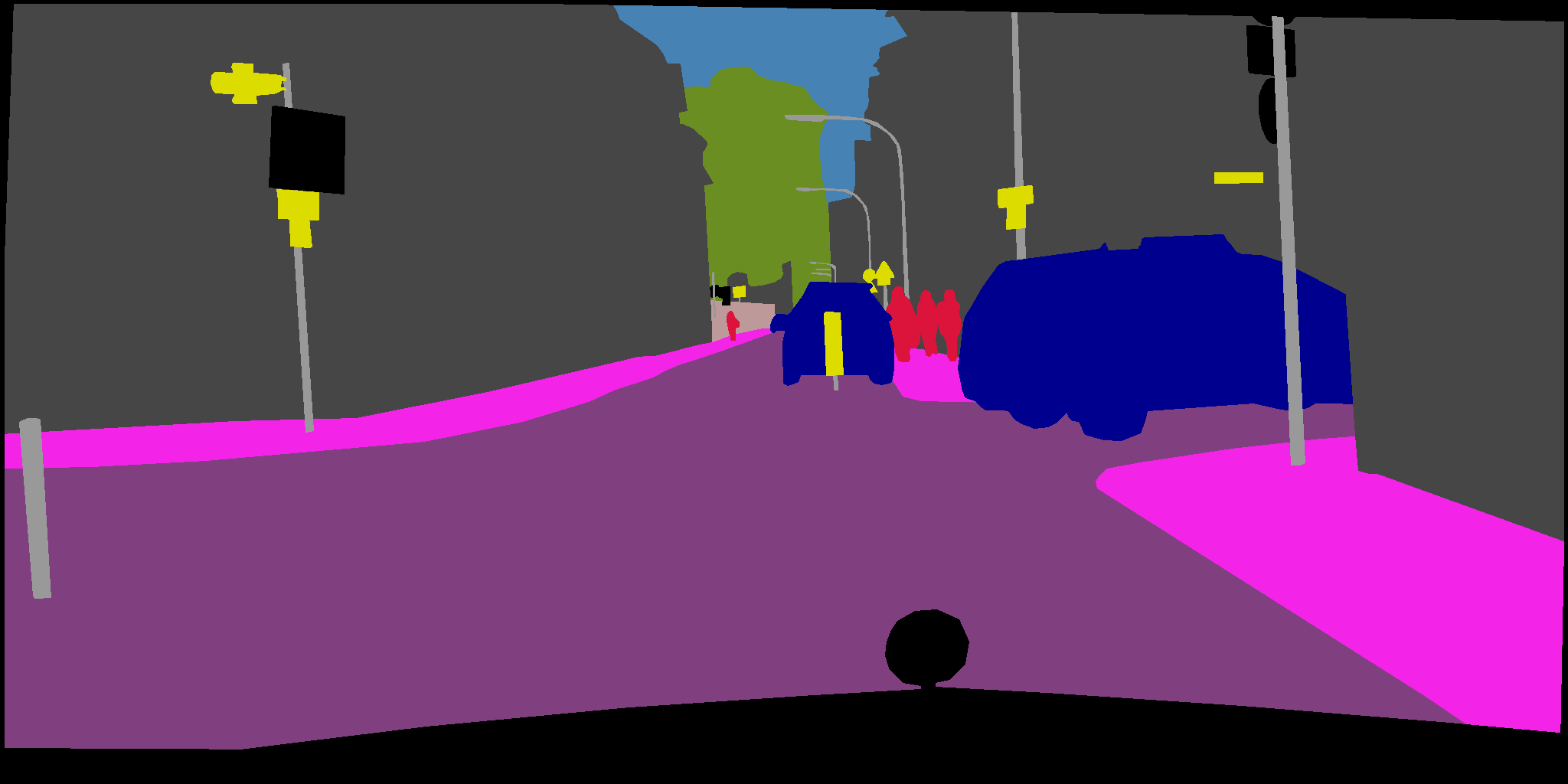}} &
\raisebox{-0.5\height}{\includegraphics[width=0.16\textwidth]{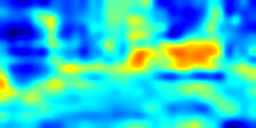}} &
\raisebox{-0.5\height}{\includegraphics[width=0.16\textwidth]{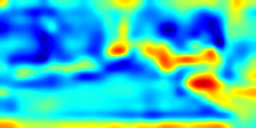}}&
\raisebox{-0.5\height}{\includegraphics[width=0.16\textwidth]{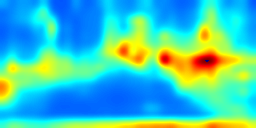}} &
\raisebox{-0.5\height}{\includegraphics[width=0.16\textwidth]{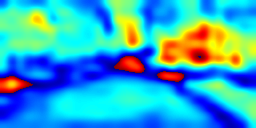}}\\
\raisebox{-0.5\height}{\includegraphics[width=0.16\textwidth]{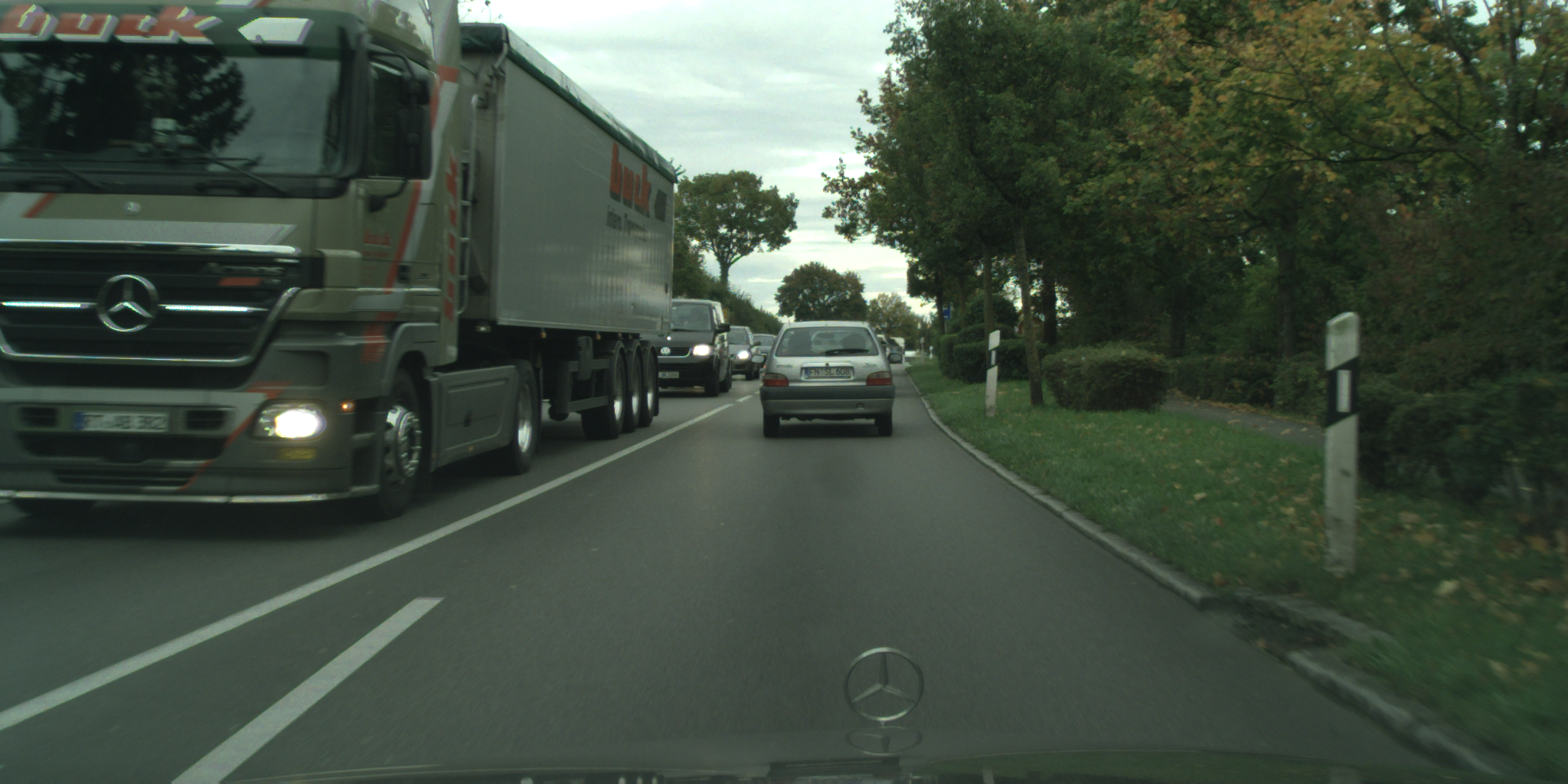}} &
\raisebox{-0.5\height}{\includegraphics[width=0.16\textwidth]{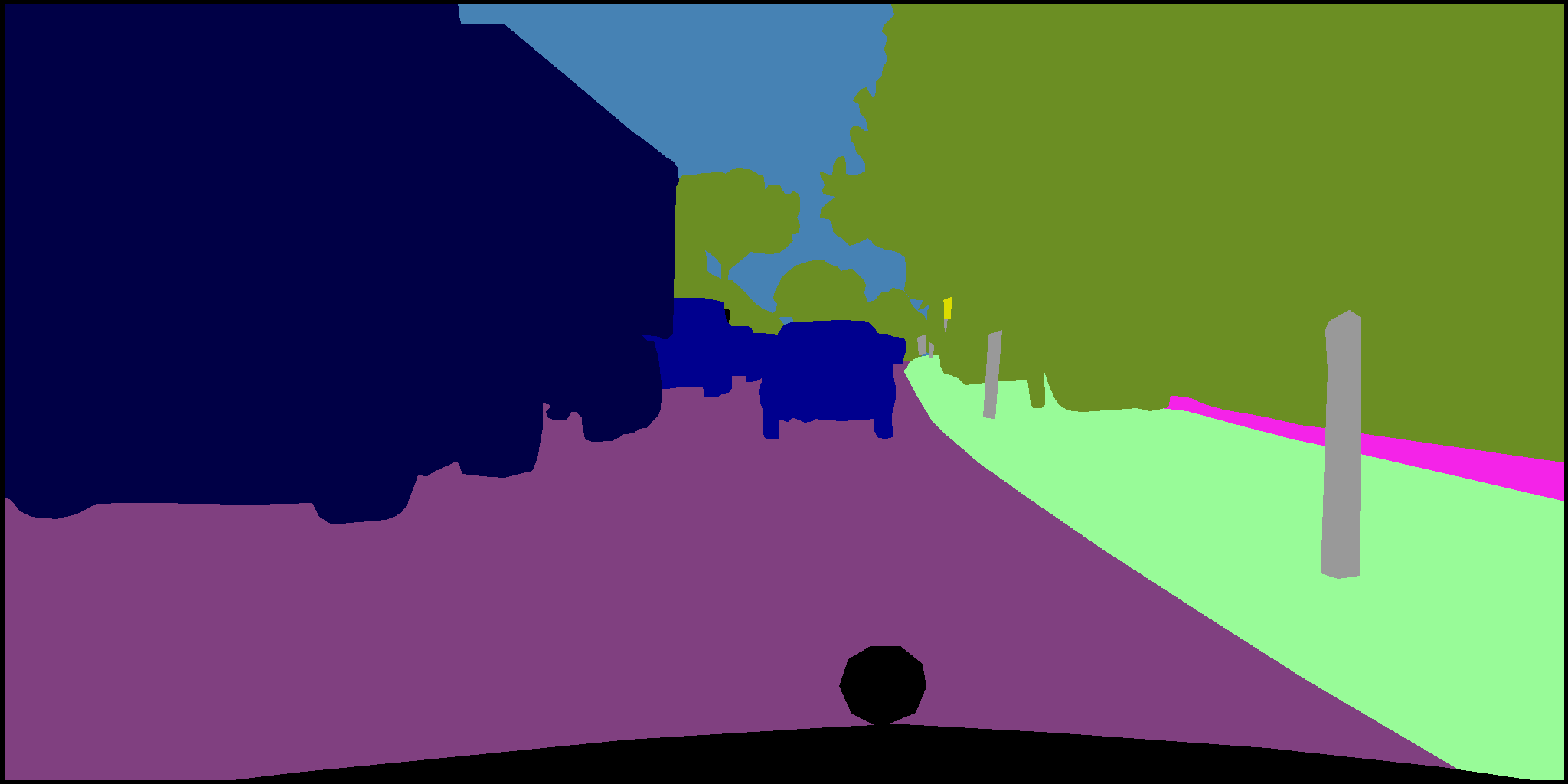}} &
\raisebox{-0.5\height}{\includegraphics[width=0.16\textwidth]{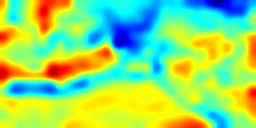}} &
\raisebox{-0.5\height}{\includegraphics[width=0.16\textwidth]{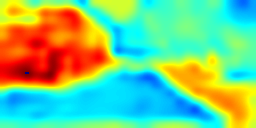}}&
\raisebox{-0.5\height}{\includegraphics[width=0.16\textwidth]{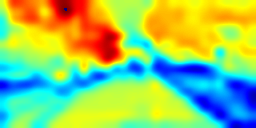}} &
\raisebox{-0.5\height}{\includegraphics[width=0.16\textwidth]{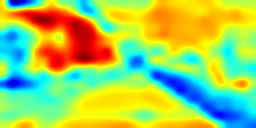}}\\
\raisebox{-0.5\height}{\includegraphics[width=0.16\textwidth]{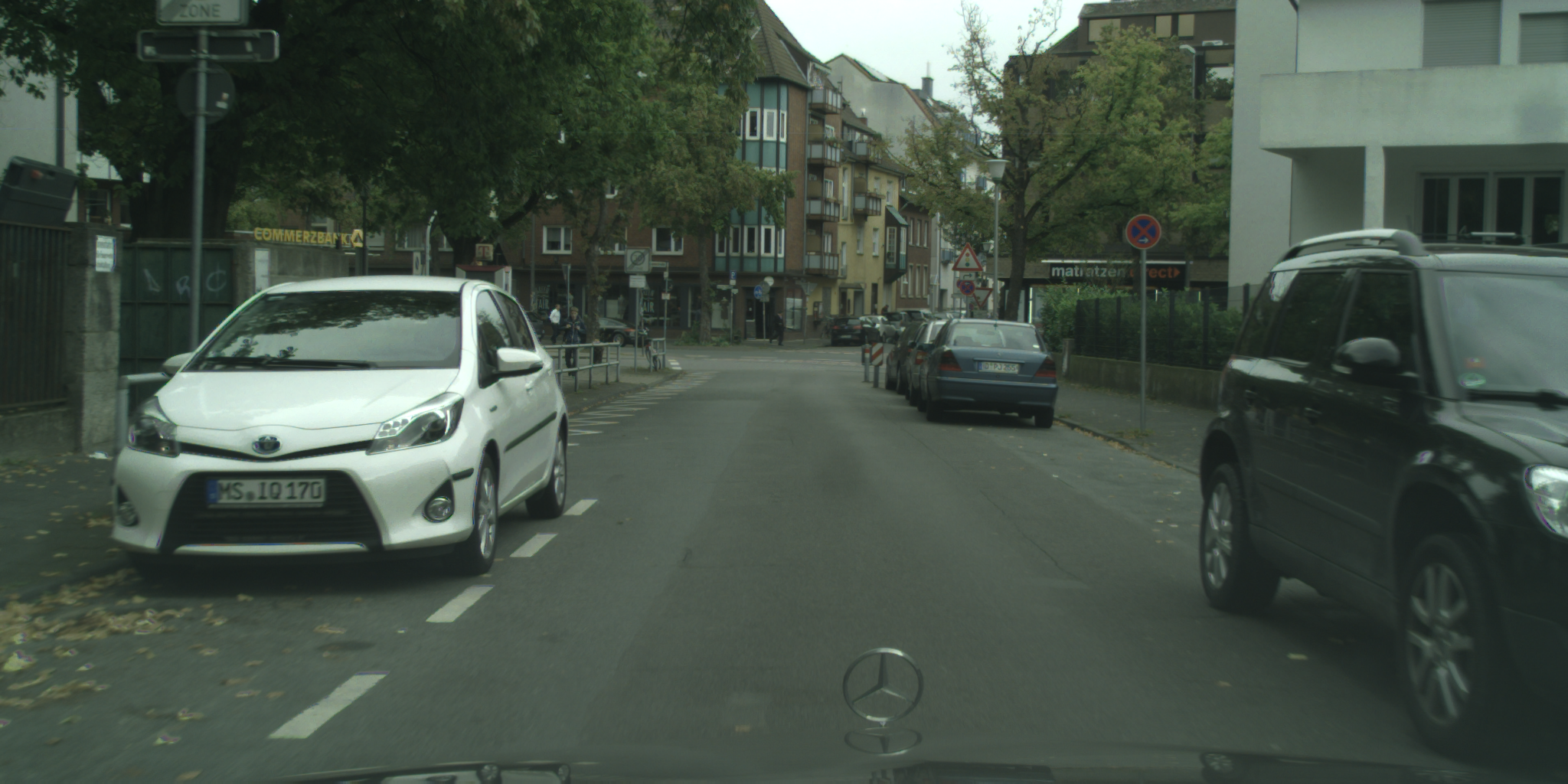}} &
\raisebox{-0.5\height}{\includegraphics[width=0.16\textwidth]{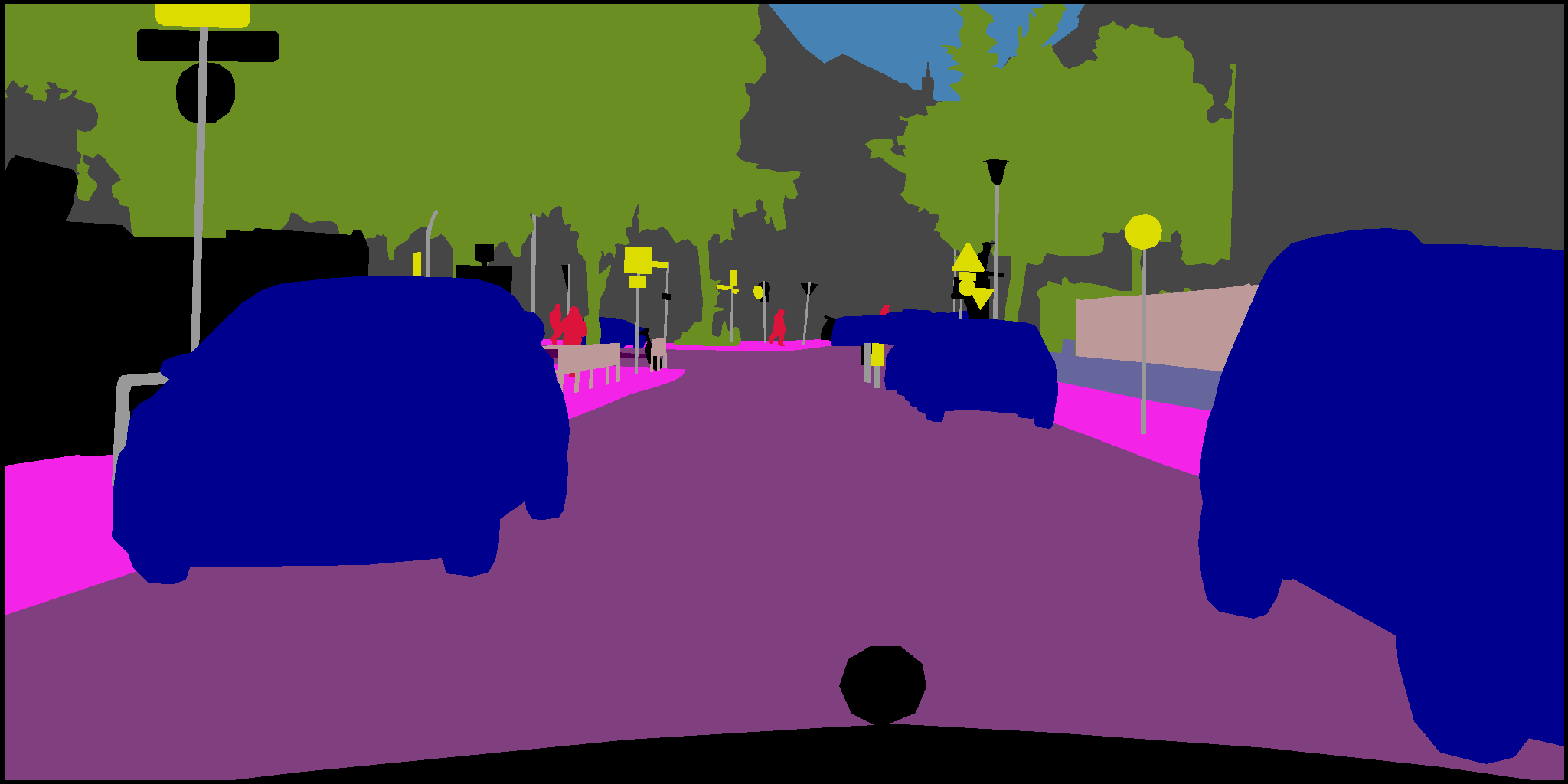}} &
\raisebox{-0.5\height}{\includegraphics[width=0.16\textwidth]{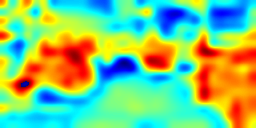}} &
\raisebox{-0.5\height}{\includegraphics[width=0.16\textwidth]{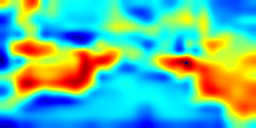}}&
\raisebox{-0.5\height}{\includegraphics[width=0.16\textwidth]{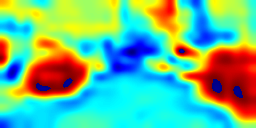}} &
\raisebox{-0.5\height}{\includegraphics[width=0.16\textwidth]{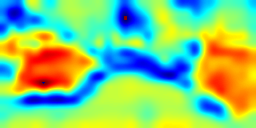}}\\
Image&Ground Truth&Student&TransKD-Base&TransKD-GL&TransKD-EA
\end{tabular}
\caption{Feature map visualizations. The feature maps at the last transformer encoder stage, \textit{i.e.}, the 4th stage, are visualized for analysis.  
}
\label{fig:feature_map_visualization}
\end{figure*}

\begin{figure*}[t!]
% \footnotesize
\small
% \scriptsize
\setlength\tabcolsep{1pt}
\centering

\begin{tabular}{c c c c c c}
\rotatebox[origin=c]{90}{Image} &
\raisebox{-0.5\height}{\includegraphics[width=0.193\textwidth]{images/cityscapes/img2.png}} &
\raisebox{-0.5\height}{\includegraphics[width=0.172\textwidth]{images/ACDC/img_fog.png}} &
\raisebox{-0.5\height}{\includegraphics[width=0.172\textwidth]{images/ACDC/img_night.png}} &
\raisebox{-0.5\height}{\includegraphics[width=0.172\textwidth]{images/ACDC/img_rain.png}} &
\raisebox{-0.5\height}{\includegraphics[width=0.172\textwidth]{images/ACDC/img_snow.png}}\\
\rotatebox[origin=c]{90}{\makecell{Ground\\Truth}} &
\raisebox{-0.5\height}{\includegraphics[width=0.193\textwidth]{images/cityscapes/gt2.png}} &
\raisebox{-0.5\height}{\includegraphics[width=0.172\textwidth]{images/ACDC/gt_fog.png}} &
\raisebox{-0.5\height}{\includegraphics[width=0.172\textwidth]{images/ACDC/gt_night.png}} &
\raisebox{-0.5\height}{\includegraphics[width=0.172\textwidth]{images/ACDC/gt_rain.png}} &
\raisebox{-0.5\height}{\includegraphics[width=0.172\textwidth]{images/ACDC/gt_snow.png}}\\
\rotatebox[origin=c]{90}{Teacher} &
\raisebox{-0.5\height}{\includegraphics[width=0.193\textwidth]{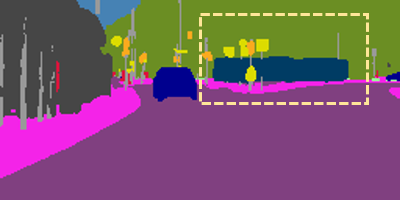}} &
\raisebox{-0.5\height}{\includegraphics[width=0.172\textwidth]{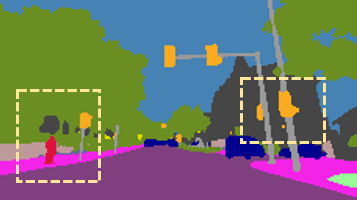}} &
\raisebox{-0.5\height}{\includegraphics[width=0.172\textwidth]{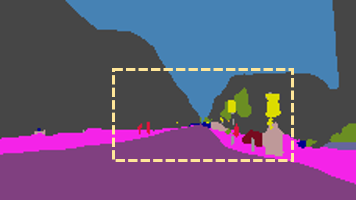}} &
\raisebox{-0.5\height}{\includegraphics[width=0.172\textwidth]{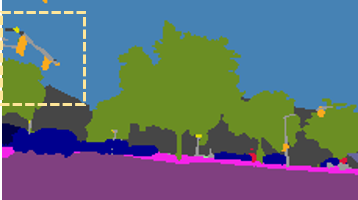}} &
\raisebox{-0.5\height}{\includegraphics[width=0.172\textwidth]{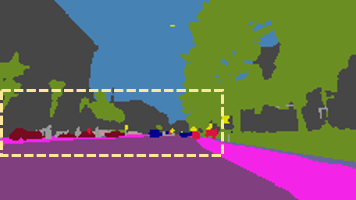}}\\
\rotatebox[origin=c]{90}{Student} &
\raisebox{-0.5\height}{\includegraphics[width=0.193\textwidth]{images/cityscapes/student2.png}} &
\raisebox{-0.5\height}{\includegraphics[width=0.172\textwidth]{images/ACDC/student_fog.png}} &
\raisebox{-0.5\height}{\includegraphics[width=0.172\textwidth]{images/ACDC/student_night.png}} &
\raisebox{-0.5\height}{\includegraphics[width=0.172\textwidth]{images/ACDC/student_rain.png}} &
\raisebox{-0.5\height}{\includegraphics[width=0.172\textwidth]{images/ACDC/student_snow.png}}\\
\rotatebox[origin=c]{90}{CD} & 
\raisebox{-0.5\height}{\includegraphics[width=0.193\textwidth]{images/cityscapes/cd2.png}} &
\raisebox{-0.5\height}{\includegraphics[width=0.172\textwidth]{images/ACDC/cd_fog.png}} &
\raisebox{-0.5\height}{\includegraphics[width=0.172\textwidth]{images/ACDC/cd_night.png}} &
\raisebox{-0.5\height}{\includegraphics[width=0.172\textwidth]{images/ACDC/cd_rain.png}} &
\raisebox{-0.5\height}{\includegraphics[width=0.172\textwidth]{images/ACDC/cd_snow.png}}\\
\rotatebox[origin=c]{90}{KR} & 
\raisebox{-0.5\height}{\includegraphics[width=0.193\textwidth]{images/cityscapes/kr2.png}} &
\raisebox{-0.5\height}{\includegraphics[width=0.172\textwidth]{images/ACDC/kr_fog.png}} &
\raisebox{-0.5\height}{\includegraphics[width=0.172\textwidth]{images/ACDC/kr_night.png}} &
\raisebox{-0.5\height}{\includegraphics[width=0.172\textwidth]{images/ACDC/kr_rain.png}} &
\raisebox{-0.5\height}{\includegraphics[width=0.172\textwidth]{images/ACDC/kr_snow.png}}\\
\rotatebox[origin=c]{90}{\makecell{TransKD\\-Base}} &
\raisebox{-0.5\height}{\includegraphics[width=0.193\textwidth]{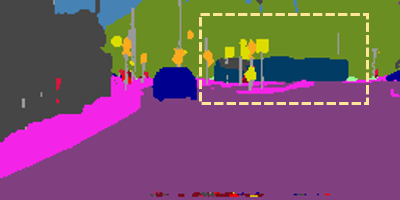}} &
\raisebox{-0.5\height}{\includegraphics[width=0.172\textwidth]{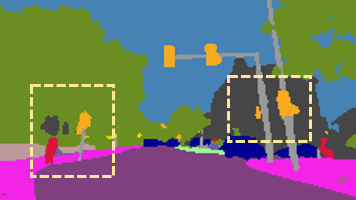}} &
\raisebox{-0.5\height}{\includegraphics[width=0.172\textwidth]{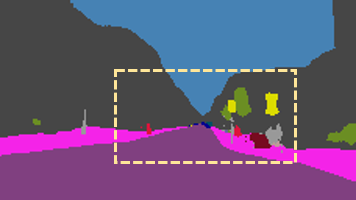}} &
\raisebox{-0.5\height}{\includegraphics[width=0.172\textwidth]{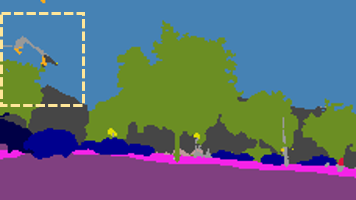}} &
\raisebox{-0.5\height}{\includegraphics[width=0.172\textwidth]{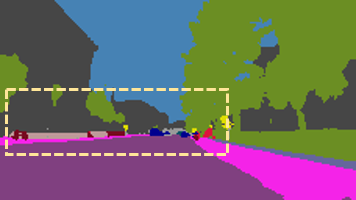}}\\
\rotatebox[origin=c]{90}{\makecell{TransKD\\-GL}} &
\raisebox{-0.5\height}{\includegraphics[width=0.193\textwidth]{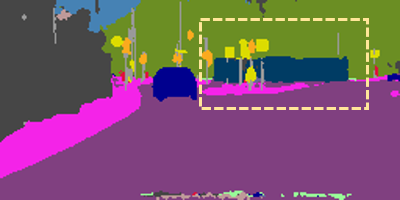}} &
\raisebox{-0.5\height}{\includegraphics[width=0.172\textwidth]{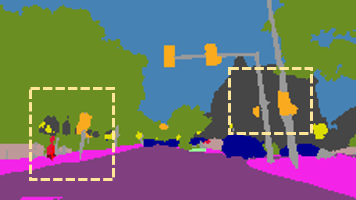}} &
\raisebox{-0.5\height}{\includegraphics[width=0.172\textwidth]{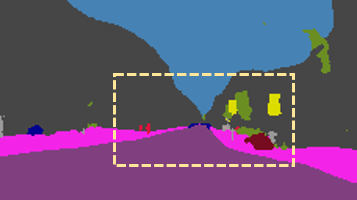}} &
\raisebox{-0.5\height}{\includegraphics[width=0.172\textwidth]{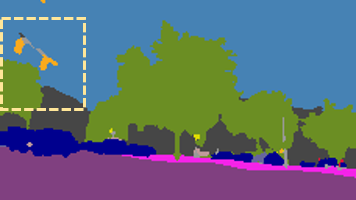}} &
\raisebox{-0.5\height}{\includegraphics[width=0.172\textwidth]{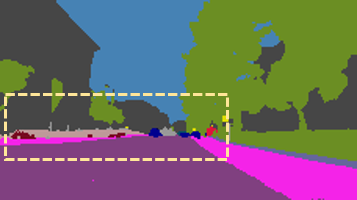}}\\
\rotatebox[origin=c]{90}{\makecell{TransKD\\-EA}} &
\raisebox{-0.5\height}{\includegraphics[width=0.193\textwidth]{images/cityscapes/transkd_ea2.png}} &
\raisebox{-0.5\height}{\includegraphics[width=0.172\textwidth]{images/ACDC/transkd_ea_fog.png}} &
\raisebox{-0.5\height}{\includegraphics[width=0.172\textwidth]{images/ACDC/transkd_ea_night.png}} &
\raisebox{-0.5\height}{\includegraphics[width=0.172\textwidth]{images/ACDC/transkd_ea_rain.png}} &
\raisebox{-0.5\height}{\includegraphics[width=0.172\textwidth]{images/ACDC/transkd_ea_snow.png}}\\
~&Normal&Fog&Night&Rain&Snow
\end{tabular}
\caption{
Qualitative street scene semantic segmentation results under normal condition from the Cityscapes dataset~\cite{cordts2016cityscapes} and adverse conditions from the ACDC dataset~\cite{sakaridis2021acdc}. 
The performances of our knowledge distillation frameworks, TransKDs, are compared with Channel-wise Knowledge Distillation (CD)~\cite{shu2021channel_distillation} and Knowledge Review (KR)~\cite{chen2021knowledge_review}.
}
\label{fig:segmentation_map_visualization_supp}
\end{figure*}
\section{More Qualitative Analyses}

\noindent\textbf{Analyses of intermediate feature maps.}
We provide visual explanations of the network's internal decision processes by visualizing the feature maps coming from the 4th stage of the model in Fig.~\ref{fig:feature_map_visualization}, with representative Cityscapes validation set examples.
Compared with the student without any TransKD components, TransKD-enhanced models present semantically recognizable responses.
TransKD helps to pay more attention on the \emph{car} and \emph{truck} objects, as shown in the reddish regions on the feature maps, while maintaining clear boundaries with the background triggering much weaker feature map responses (blue and green regions),
leading to better semantic segmentation outcome.
Apart from that, TransKD-Base tends to focus on large objects, \textit{e.g.}, on the \emph{truck}, while TransKD-GL shows more evident responses on \emph{cars} which have relatively smaller size, which we attribute to the extraction of both local and global cues encouraged by this approach.
Moreover, TransKD-EA better spotlights the objects which are far away from the camera, \textit{e.g.}, the \emph{car} in the first row.
These visualization results further illustrate the effectiveness of TransKD for learning semantic-aware intermediate representations.

\noindent\textbf{Analyses of outdoor semantic segmentation.}
Here we compare the segmentation outputs of the most effective response-based distillation framework CD~\cite{shu2021channel_distillation}, feature-based distillation framework KR~\cite{chen2021knowledge_review}, and all three versions of TransKD.
The outdoor semantic segmentation experiments are conducted on Cityscapes~\cite{cordts2016cityscapes} (normal scenes) and ACDC dataset~\cite{sakaridis2021acdc} (adverse scenes).
According to the quantitative results, TransKD-GL shows the most satisfactory performance on the \emph{snowy} weather, while TransKD-EA outperforms the all counterparts on normal and other adverse weathers.
As shown in Fig.~\ref{fig:segmentation_map_visualization_supp}, the student SegFormer B0 without pre-training, and distilled with CD and KR, can roughly recognize the scenes, such as \emph{road}, \emph{sidewalk}, and \emph{vegetation}, but the compact student transformer is still unable to segment the details, \textit{e.g.}, the \emph{traffic light} and \emph{pedestrian} in the distance and the occluded \emph{bus}.
On the contrary, TransKDs extracts both spatial and sequential relations through the combination of feature map and patch embedding distillation, thereby improving the consistency of the \emph{bus}.
Moreover, TransKDs can effectively distinguish the \emph{people}, \emph{rider}, and \emph{bicycle} in adverse weathers, even though they are far away.

\begin{figure}[t!]
% \footnotesize
\small
% \scriptsize
\setlength\tabcolsep{1pt}
\centering

\begin{tabular}{c c c}
\raisebox{-0.5\height}{\includegraphics[width=0.16\textwidth]{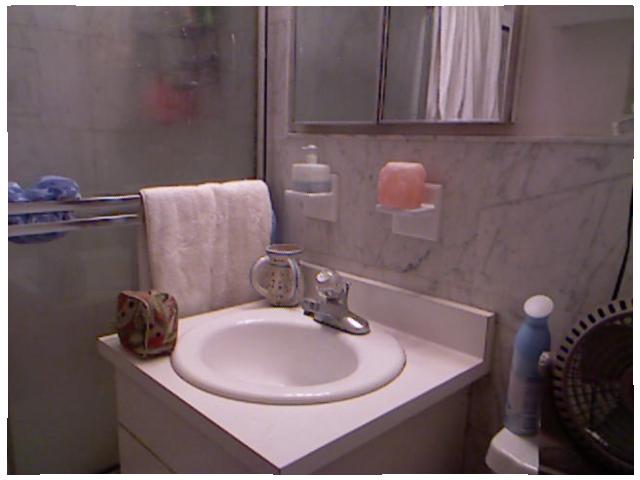}}&
\raisebox{-0.5\height}{\includegraphics[width=0.16\textwidth]{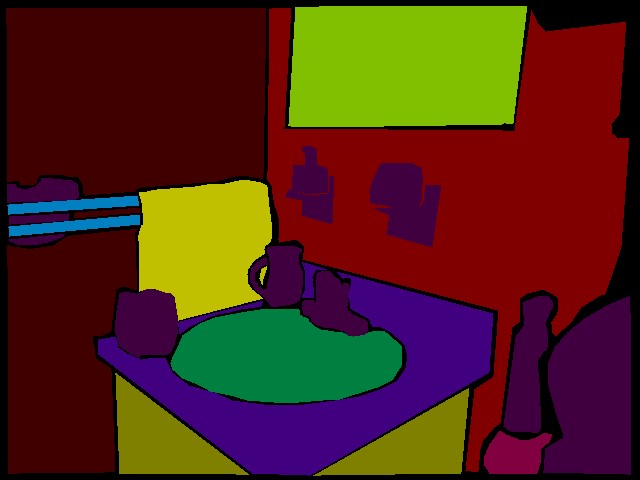}} &
\raisebox{-0.5\height}{\includegraphics[width=0.16\textwidth]{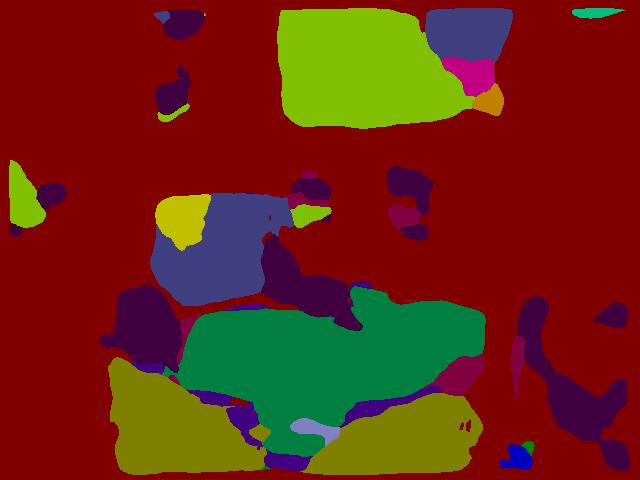}}\\
\raisebox{-0.5\height}{\includegraphics[width=0.16\textwidth]{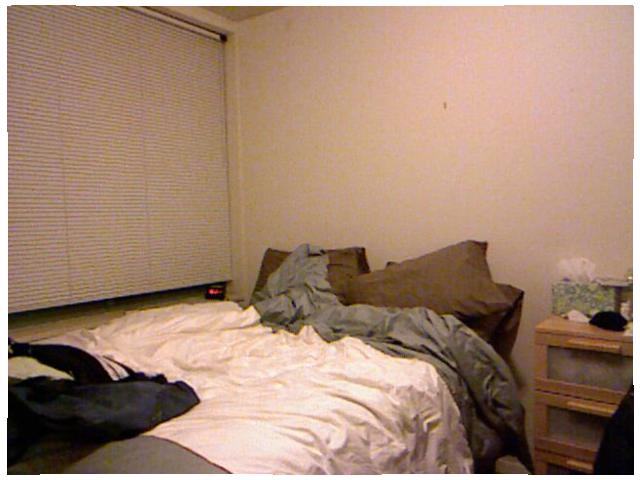}}&
\raisebox{-0.5\height}{\includegraphics[width=0.16\textwidth]{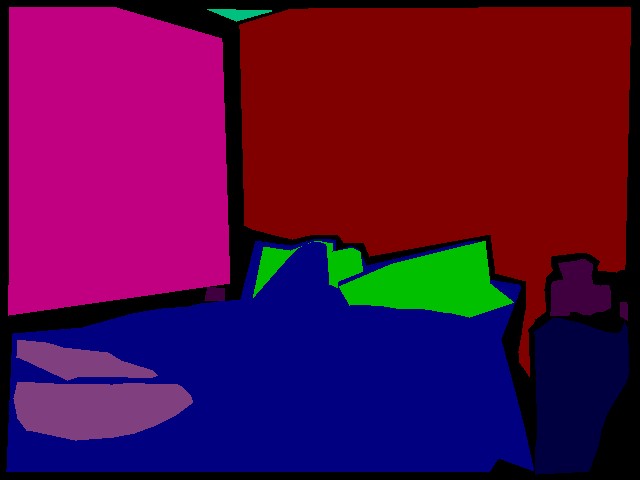}} &
\raisebox{-0.5\height}{\includegraphics[width=0.16\textwidth]{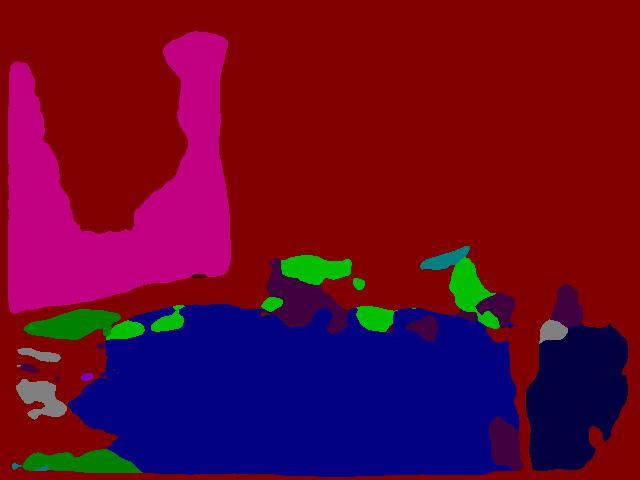}}\\
Image&Ground Truth&Student\\
\raisebox{-0.5\height}{\includegraphics[width=0.16\textwidth]{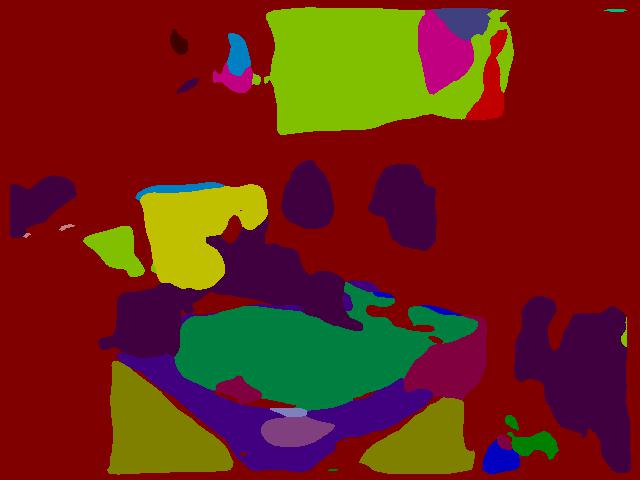}}&
\raisebox{-0.5\height}{\includegraphics[width=0.16\textwidth]{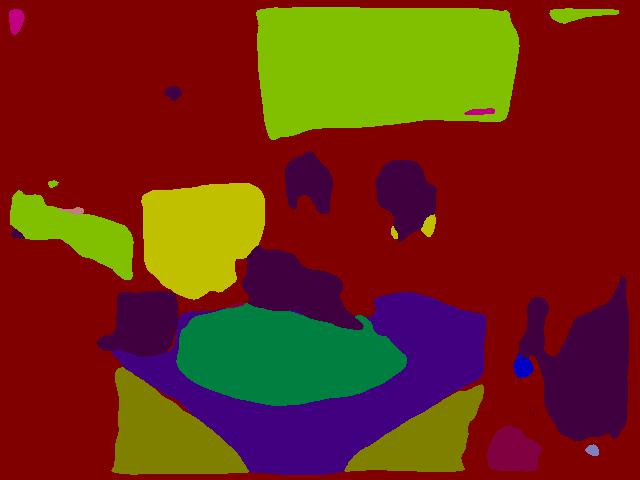}} &
\raisebox{-0.5\height}{\includegraphics[width=0.16\textwidth]{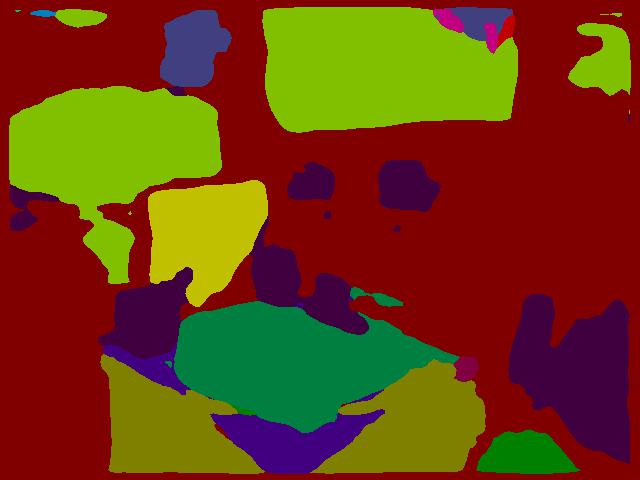}}\\
\raisebox{-0.5\height}{\includegraphics[width=0.16\textwidth]{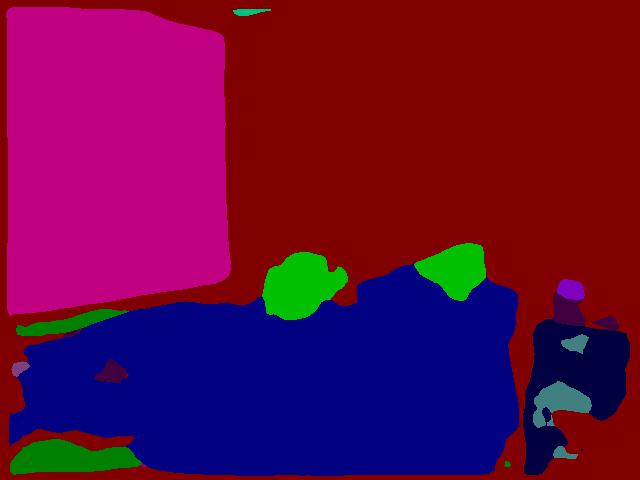}}&
\raisebox{-0.5\height}{\includegraphics[width=0.16\textwidth]{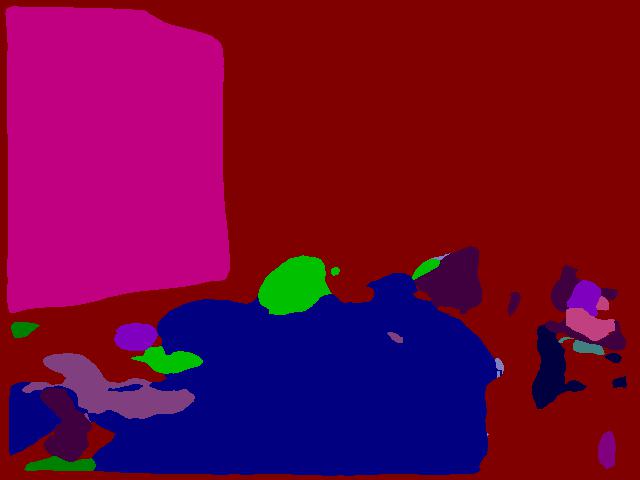}} &
\raisebox{-0.5\height}{\includegraphics[width=0.16\textwidth]{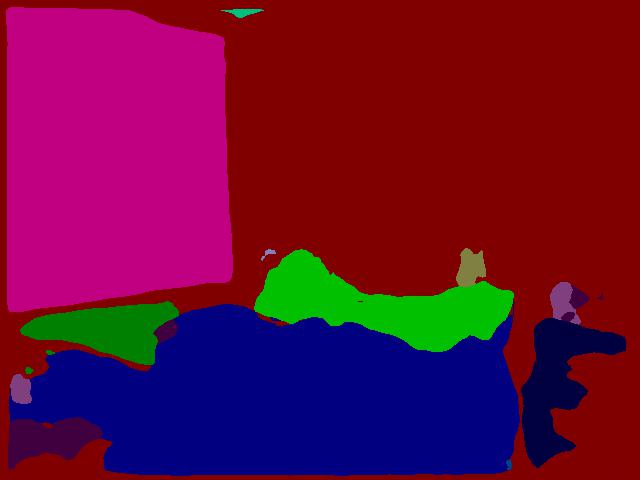}}\\
TransKD-Base&TransKD-GL&TransKD-EA
\end{tabular}
\caption{Qualitative indoor scene semantic segmentation results of our TransKD methods on the NYUv2 dataset~\cite{silberman2012nyu_dataset}.}
\label{fig:nyuv2_visualization}
\end{figure}

\noindent\textbf{Analyses of indoor semantic segmentation.}
In our final study, we turn to the \textit{indoor} semantic segmentation results and illustrate the segmentation outcomes on the NYUv2 dataset~\cite{silberman2012nyu_dataset} in Fig.~\ref{fig:nyuv2_visualization}.
Compared with the segmentation results produced by the student without any distillation, the segmentation boundaries of TransKD are much clearer, and most of the details of the indoor scenes are preserved, especially for the bedroom scene, \textit{e.g.}, on \emph{pillow}. For the bathroom example, similar results can be found, where, for example, \emph{towel} is better recognized by our approaches. 
Apart from the objects with relatively small size, objects with less texture cues can be completely segmented, \textit{e.g.}, the \emph{wall} in the bedroom.
These qualitative examples consistently confirm good generalization of TransKD across diverse environments (indoor and outdoor, different weather conditions, \textit{etc.}) as well as the benefits of knowledge transfer from both patch embeddings and feature maps in transformer-to-transformer knowledge distillation.
